%% file: v5.tex
\documentclass[12pt]{article}
\usepackage{amssymb, amsmath, adjustbox, amsthm, mathtools, relsize, bbm, algorithmic, algorithm, microtype, graphicx, booktabs, color, multirow, rotating, latexsym, import, apacite, subfig}
\usepackage[authoryear]{natbib}
\usepackage[title]{appendix}

\usepackage{times, url}

\subimport{.}{_math}

\usepackage{hyperref}
\newcommand{\captionfonts}{\normalsize}

\makeatletter  
\long\def\@makecaption#1#2{%
  \vskip\abovecaptionskip
  \sbox\@tempboxa{{\captionfonts #1: #2}}%
  \ifdim \wd\@tempboxa >\hsize
    {\captionfonts #1: #2\par}
  \else
    \hbox to\hsize{\hfil\box\@tempboxa\hfil}%
  \fi
  \vskip\belowcaptionskip}
\makeatother   

\begin{document}
\hspace{13.9cm}1

\ \vspace{20mm}\\
{\LARGE On Kernel Method-Based Connectionist Models and Supervised Deep Learning Without Backpropagation 
}

\ \\
{\bf \large Shiyu Duan$^{\displaystyle 1}$}\\
{\bf \large Shujian Yu$^{\displaystyle 1}$}\\
{\bf \large Yunmei Chen$^{\displaystyle 2}$}\\
{\bf \large Jose C.~Principe$^{\displaystyle 1}$}\\
{$^{\displaystyle 1}$Department of Electrical and Computer Engineering, University of Florida}\\
{$^{\displaystyle 2}$Department of Mathematics, University of Florida}\\
%


\thispagestyle{empty}
\markboth{}{NC instructions}
\ \vspace{-0mm}\\
%
\begin{center} {\bf Abstract} \end{center}
We propose a novel family of connectionist models based on kernel machines and consider the problem of learning layer-by-layer a compositional hypothesis class, i.e., a feedforward, multilayer architecture, in a supervised setting. In terms of the models, we present a principled method to ``kernelize'' (partly or completely) any neural network (NN). With this method, we obtain a counterpart of any given NN that is powered by kernel machines instead of neurons. In terms of learning, when learning a feedforward deep architecture in a supervised setting, one needs to train all the components simultaneously using backpropagation (BP) since there are no explicit targets for the hidden layers~\citep{Rumelhart86}. We consider without loss of generality the two-layer case and present a general framework that explicitly characterizes a target for the hidden layer that is optimal for minimizing the objective function of the network. This characterization then makes possible a purely greedy training scheme that learns one layer at a time, starting from the input layer. We provide realizations of the abstract framework under certain architectures and objective functions. Based on these realizations, we present a layer-wise training algorithm for an $l$-layer feedforward network for classification, where $l\geq 2$ can be arbitrary. This algorithm can be given an intuitive geometric interpretation that makes the learning dynamics transparent. Empirical results are provided to complement our theory. We show that the kernelized networks, trained layer-wise, compare favorably with classical kernel machines as well as other connectionist models trained by BP. We also visualize the inner workings of the greedy kernelized models to validate our claim on the transparency of the layer-wise algorithm.


\section{Introduction}
One can ``kernelize'' any neural network (NN) by replacing each artificial neuron \citep{Mcculloch1943}, i.e., function approximator of the form $\fnwa{f}{\vctr{x}} = \sigma\paren{\dotpd{\vctr{w}}{\vctr{x}} + b}$, with a kernel machine, i.e., function approximator of the form $\fnwa{f}{\vctr{x}} = \innersub{\vctr{w}}{\vfnwa{\phi}{\vctr{x}}}{H} + b$ with kernel function $\kernelwa{\vctr{x}}{\vctr{y}} = \innersub{\vfnwa{\phi}{\vctr{x}}}{\vfnwa{\phi}{\vctr{y}}}{H}$. While the nonlinearities in deep NNs make it notoriously difficult to analyze these models, the simple interpretation of a kernel machine as a hyperplane in a reproducing kernel Hilbert space (RKHS) makes the kernelized networks more tractable mathematically. We shall refer to the kernelized NNs in general as kernel networks (KNs). 

We then revisit the problem of learning a composite hypothesis class, by which we mean a trainable model that consists of more elementary trainable submodels, in a supervised learning setting. In this paper, we shall only consider the special case of a compositional hypothesis class, in which the elementary submodels are linked via function compositions and therefore the overall model can be written as $\vfn{F} = \vfnsub{F}{l}\circ\cdots\circ \vfnsub{F}{1}$ for some $l$, with each $\vfnsub{F}{i}$ being a submodel with proper domain and codomain. For example, a deep, feedforward NN can be considered as a compositional hypothesis class. 

When it comes to training these models, the usual method is to learn all its trainable submodels simultaneously using, for example, backpropagation (BP)~\citep{Rumelhart86}. However, in the context of supervised learning, the need for BP is caused by the fact that there is no explicit target information to tune the latent submodels \citep{Rumelhart86}. Moreover, when the model is large, BP usually becomes computationally intensive and can suffer from issues such as vanishing gradient. Also, BP returns very little information on the training of each submodel to the user and therefore forces the user to treat the model as a ``black box''. For example, it is usually not possible to know which specific part or parts of the network is responsible when the performance is suboptimal. Also, it is extremely difficult to interpret or assess the hidden representations during or after training.

We consider the problem of reducing the compositional learning problem into a set of noncompositional ones and then solving each one of them individually. We approach by deriving explicit targets for the hidden submodels. The targets are optimal for minimizing a given objective function of the overall model. The central idea can be summarized as follows: let input data $S_\vctr{X}$, supervision $S_Y$ (labels in classification, dependent variable in regression), a two-layer feedforward architecture $\vfnsub{F}{2}\circ \vfnsub{F}{1}$, and an objective function $\fnwa{\tilde{R}}{\vfnsub{F}{2}\circ \vfnwasub{F}{S_\vctr{X}}{1}\comma S_Y}$ be given, define $\vfndub{F}{2}{\star}\circ \vfndub{F}{1}{\star} \coloneqq \argmin_{\vfnsub{F}{2}\circ \vfnsub{F}{1}} \fnwa{\tilde{R}}{\vfnsub{F}{2}\circ \vfnwasub{F}{S_\vctr{X}}{1}\comma S_Y}$. If we could find functions $s\cm u$, and a new objective $\tilde{R}_1(\fnwa{s}{\vfnwasub{F}{S_\vctr{X}}{1}}\cm u(S_Y))$ whose minimizer is equivalent to $\vfndub{F}{1}{\star}$ for minimizing the objective $\tilde{R}$, then finding $\vfndub{F}{1}{\star}$ is equivalent to finding an $\vfnsub{F}{1}$ that minimizes $\tilde{R}_1$. If the dependence of $s$ and $u$ on $\vfnsub{F}{2}$ can be reduced to a point where this search for $\vfndub{F}{1}{\star}$ does not involve the trainable parameters of $\vfnsub{F}{2}$, then we have reduced the original compositional learning problem into two noncompositional ones that can be solved sequentially.

As examples, we provide realizations of the abstract framework and also, based on these realizations, a sample greedy training algorithm for a multilayer feedforward architecture for classification. This greedy learning algorithm enjoys the same optimality guarantee as BP in the sense that they both effectively train each layer to minimize the overall objective. But the former is faster, more memory efficient, and evidently less susceptible to vanishing gradient. It also greatly increases the transparency of deep models: the quality of learning in the hidden layers can be directly assessed during or after training, providing the user with more information about training. Also, alternative model selection and hyperparameter tuning paradigms are now available since unsatisfying performance of the network can be traced to a certain layer or layers, allowing the user to ``debug'' the layers individually. Moreover, the target for each hidden layer in this algorithm can be given an intuitive geometric interpretation, making the learning dynamics transparent. 


Empirical results are provided to complement our theory. First, we compare KNs with classical kernel machines and show that KNs consistently outperform Support Vector Machine (SVM) \citep{Cortes1995} as well as several SVMs enhanced by Multiple Kernel Learning (MKL) algorithms \citep{Bach04, Gonen11}. We then fully or partly kernelized both fully-connected and convolutional NNs and trained them with the proposed layer-wise algorithm. The resulting KNs compare favorably with their NN equivalents trained with BP as well as some other commonly-used deep architectures trained with BP together with unsupervised greedy pre-training. We also visualize the learning dynamics and hidden representations in the greedy kernelized networks to validate our claim on the transparency of the greedy algorithm.

\section{Setting and Notations}
\label{assumptions and notations}

We consider the following supervised set-up: let a realization of an i.i.d. random sample be given: $S = \seqdub{\paren{\vctrsub{x}{n}\comma y_n}}{n=1}{N}$, where $\paren{\vctrsub{x}{n}\comma y_n} \in \spcsup{R}{d_0} \times \spc{R}$. Denote $\seqdub{\vctrsub{x}{n}}{n=1}{N}$ as $S_{\rvctr{X}}$ and $\seqdub{y_n}{n=1}{N}$ as $S_Y$ for convenience. We consider only real, continuous, symmetric, positive definite (PD) kernels \citep{Scholkopf01}, which possess the reproducing property $\kernelwa{\vctr{x}}{\vctr{y}} = \innersub{
\fnwa{\phi}{\vctr{x}}}{
	\fnwa{\phi}{\vctr{y}}
}{H}$, where $H$ is the RKHS induced by $k$. Further, we assume, for all kernels considered in all results, that $\kernelwa{\vctr{x}}{\vctr{x}} = c < +\infty\cm\forall\vctr{x}$, and that $\inf_{\vctr{x}, \vctr{y}} \kernelwa{\vctr{x}}{\vctr{y}} = a > -\infty$. It is straightforward to check using Cauchy-Schwarz inequality that the first condition implies $\max_{\vctr{x}, \vctr{y}} \kernelwa{\vctr{x}}{\vctr{y}} = c$. Note that by construction of a PD kernel, we always have $a<c$.

For the rest of this paper, we shall use bold letters to denote vectors or vector-valued functions. For random elements, we use capital letter to denote the random element and lower-case letter a realization of it. Notations similar to the following will be used whenever convenient: for a general $l$-layer feedforward architecture $\vfnsub{F}{l} \circ \cdots \circ \vfnsub{F}{1}$ and for $i =2, 3, \ldots, l\cm \vctr{x}\in\spcsup{R}{d_0}$, $\vfnwasub{F}{\vctr{x}}{i}\coloneqq \vfnsub{F}{i} \circ \cdots \circ \vfnwasub{F}{\vctr{x}}{1}$. For any $\vfn{F}$, the shorthand $\vfnwa{F}{S_{\rvctr{X}}}$ represents $\seqdub{\vfnwa{F}{\vctrsub{x}{n}}}{n=1}{N}$. And likewise for $\vfnwa{F}{S_Y}$. When there is no confusion, we shall suppress the dependency of any loss function on the example for brevity, i.e., for a loss function $\ell$, instead of writing $\fnwa{\ell}{\fnwa{f}{\vctr{x}},\, y}$, we write $\fnwa{\ell}{f}$. 

Given a loss function $\fnwa{\ell}{\fnwa{f}{\vctr{x}}\cm y}$, we define the risk as $R(f) \coloneqq \E_{\paren{\vctr{X}\cm Y}}\fnwa{\ell}{\fnwa{f}{\vctr{X}}\cm Y}$ and an objective function $\fnwa{\tilde{R}}{\fnwa{f}{S_\vctr{X}}\cm S_Y}$ to be a bound on the risk that is computable using the given data only. In this paper, we shall take any objective as given without rigorously justifying why it is a bound of some risk since that is not the purpose of this paper. Nevertheless, the objectives we use in this paper are fairly common and the corresponding justifications are routine. We make this distinction between risk and objective here as it will be needed in later discussions.

\section{Kernelizing a Neural Network}
\label{defining kmlp}

Kernel machines are parametric models defined as $\fnwa{f}{\vctr{x}} = \innersub{\vctr{w}}{\vfnwa{\phi}{\vctr{x}}}{H} + b$ with kernel $\kernelwa{\vctr{x}}{\vctr{y}} = \innersub{\vfnwa{\phi}{\vctr{x}}}{\vfnwa{\phi}{\vctr{y}}}{H}$ and $\vctr{w}\cm b$ being the learnable weights and $\phi$ being a map into the RKHS $H$. NNs are connectionist models defined by arbitrarily combining the parametric base units defined as $\fnwa{f}{\vctr{x}} = \sigma\paren{\dotpd{\vctr{w}}{\vctr{x}} + b}$ with $\vctr{w}\cm b$ being the learnable weights and $\sigma$ a (usually nonlinear) gating function. These base units are sometimes called neurons.

While NNs are flexible models and have strong expressive power in practice, they are notoriously difficult to analyze due to each nonlinear neuron being a nontrivial function itself and the arbitrariness involved in the overall architecture design. Kernel machines, in comparison, are much more mathematically tractable since they are linear models in the feature space $H$, i.e., the $f$ is linear in $\vctr{w}$. This allows one to reduce otherwise abstract problems into geometric ones, making possible simpler and more intuitive solutions.  However, their architectures are not as flexible and their practical performance in most cutting-edge machine learning applications has been unsatisfying~\citep{Bengio2013}. 

The question we consider is how to combine the idea of connectionism, which is central to NNs, with kernel machines and build families of models that are flexible, expressive, and at the same time, more mathematically tractable than NNs. We hope this will be a first step toward explaining why deep learning performs so well in the most challenging AI tasks. 

In this section, we discuss how to kernelize an NN to build models that combine the best of both worlds. We first present the generic approach and then as an example, concretely define a fully-kernelized Multilayer Perceptron (MLP). To further shed light on the effect of kernelization on the expressive power of the original model, we give an analysis on the model complexity of a fully-kernelized MLP. 

\subsection{A Generic Approach to Kernelization} 

\begin{figure}
\centering
\includegraphics[width=.9\linewidth]{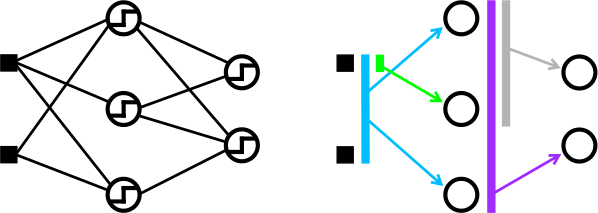}

\caption{Any NN (left, presented in the usual weight-nonlinearity abstraction) can be abstracted as a ``graph'' (right) with each node representing a neuron and each edge the input-output relationship between neurons. If a node receives multiple inputs, we view its input as a vector in some Euclidean space, as indicated by the colored rectangles. Under this abstraction, each neuron can be directly replaced by a kernel machine mapping from the same Euclidean space into the real line without altering the architecture and functionality of the model.}
\label{kernelization}
\end{figure}

The general idea we adopt is to build connectionist models with the base units being not neurons but kernel machines. This is mathematically viable since in an NN, any neuron can be directly replaced by a kernel machine without altering the architecture and functionality of the network. An illustration of this kernelization procedure is provided in Fig.~\ref{kernelization}. In this way, one can kernelize an NN to any degree: a node, several nodes, a layer, several layers, or the entire network. 

KN is flexible in the sense that one can inject prior knowledge into the architecture design, as is done for NNs. KN inherits the expressive power of the original NN since a kernel machine is a universal function approximator under mild conditions \citep{Park91, Micchelli06}. Moreover, KN works in a more mathematically intuitive way since each base unit is a simple linear model in an RKHS. 

Further, a general criticism toward kernel methods in machine learning is that their performance usually relies heavily on the parameterization of the kernels used. This issue is mitigated in KN, thanks to the introduction of connectionism. To be specific, KN performs nonparametric kernel learning alongside learning to perform the given task. Indeed, to build the network one only needs generic kernels, but in a connectionist model, the kernels on the non-input layers admit the form $\kernelwa{\vfnwa{F}{\vctr{x}}}{\vfnwa{G}{\vctr{y}}}$, where $\vfn{F}\cm\vfn{G}$ are some other trainable submodels. The fact that $\vfn{F}\cm\vfn{G}$ are trainable makes this kernel ``adaptive'', mitigating to some extent any limitation of the fixed generic kernel $k$. The training of $\vfn{F}$ and $\vfn{G}$ makes this adaptive kernel optimal as a constituent part of the corresponding kernel machine for the task the network was trained for. And it is always a valid kernel if the generic kernel $k$ is. Note that observations similar to this one have been made in different contexts by, for example, \cite{Huang2006} and \cite{Bengio2013}, we include it here only for completeness.

\subsection{Kernelized MLP: The Architecture}

As a more concrete example, we now define a fully-kernelized $l$-layer MLP, which we will specifically refer to as kernel MLP (kMLP).\footnote{A PyTorch-based \citep{Paszke2017} library for implementing KN and the proposed layer-wise training algorithm is available at: \emph{https://github.com/michaelshiyu/kerNET}.}

The $l$-layer kMLP is defined as follows. For $i\geq 1$, the $\ith{i}$ layer in a kMLP, denoted $\vfnsub{F}{i}$, is an array of $d_i$ kernel machines: $\vfnsub{F}{i}: \spcsup{R}{d_{i-1}} \to \spcsup{R}{d_i}, \vfnwasub{F}{\vctr{x}}{i} = \paren{\fnwadub{f}{\vctr{x}}{i}{1}\comma \fnwadub{f}{\vctr{x}}{i}{2}\comma \ldots\comma \fnwadub{f}{\vctr{x}}{i}{d_i}}$ with the $\fndub{f}{i}{j}$ all using kernel $k_i$. Let $\vfnsub{F}{0}$ be the identity map on $\spcsup{R}{d_0}$, each $\fndub{f}{i}{j} : \spcsup{R}{d_{i-1}} \to \spc{R}$ is a hyperplane in RKHS $H_i$: $\fnwadub{f}{\vctr{x}}{i}{j} = \innersub{\vctrsub{w}{\fndub{f}{i}{j}}}{\vfnwasub{\phi}{\vfnsub{F}{i-1}\circ \cdots \circ \vfnwasub{F}{\vctr{x}}{0}}{i}}{H_i} + b_{\fndub{f}{i}{j}}\comma \vctrsub{w}{\fndub{f}{i}{j}} \in H_i\comma b_{\fndub{f}{i}{j}} \in \spc{R}$. The set of mappings $$\setwcolon{\vfnsub{F}{l}\circ \cdots\circ \vfnsub{F}{1}}{\vctrsub{w}{\fndub{f}{i}{j}}\in H_i\cm b_{\fndub{f}{i}{j}}\in \spc{R} \text{ for all admissible $n\comma j\cm i$}}$$ defines an $l$-layer kMLP. 

In practice, $\vctrsub{w}{\fndub{f}{i}{j}}$ is usually not accessible but can be approximated using, for instance, $\sum_{n=1}^N \alpha_{i\cm n}^{j} \vfnwasub{\phi}{\vfnsub{F}{i-1}\circ \cdots \circ \vfnwasub{F}{\vctrsub{x}{n}}{0}}{i}$, where the $\alpha_{i\cm n}^{j} \in \spc{R}$ are the learnable parameters.\footnote{The optimality of this expansion can be justified in the following layer-wise setting by directly applying the representer theorem \citep{Scholkopf01a}.} 

\subsection{Kernelized MLP: Model Complexity}

We give a bound on the model complexity of an $l$-layer kMLP using a well-known complexity measure called Gaussian complexity \citep{Bartlett2002}. In particular, the bound describes the relationship between the depth/width of the model and the complexity of its hypothesis class, providing insights into the effect of kernelization on the expressive power of the model as well as useful information for model selection. We first review the definition of Gaussian complexity.

\begin{definition}[\emph{Gaussian complexity}]
Let $\rvarsub{X}{1}\comma \ldots\comma \rvarsub{X}{N}$ be i.i.d. random elements defined on metric space $\spc{X}$ and let $\spc{F}$ be a set of functions mapping from $\spc{X}$ into $\spc{R}$. Define
\begin{align*}
\egcpxwasub{\spc{F}}{N} = \Ewa{\sup_{\fn{f}\in \spc{F}} \frac{1}{N}\sum_{n=1}^N \rvarsub{Z}{n} \fnwa{f}{\rvarsub{X}{i}}\,\Bigg\vert \, \rvarsub{X}{1}\comma \ldots\comma \rvarsub{X}{N}},
\end{align*}
where $\rvarsub{Z}{1}\comma \ldots\comma \rvarsub{Z}{N}$ are independent standard normal random variables. The Gaussian complexity of $\spc{F}$ is defined as $\gcpxwasub{\spc{F}}{N} = \E \egcpxwasub{\spc{F}}{N}$.
\end{definition}
Intuitively, Gaussian complexity quantifies how well elements in a given function class can be correlated with a normally-distributed noise sequence of length $N$ \citep{Bartlett2002}.

For Proposition~\ref{gaussian complexity kmlp} and the lemma based on which this proposition is proven (Lemma~\ref{2-norm Gaussian complexity} in Appendix~\ref{proofs}), we impose the following smoothness assumption on all kernels considered: for each fixed $\vctr{x} \in \spcsup{R}{d_{i-1}}$, we assume that $\kernelwasub{\vctr{x}}{\vctr{y}}{i}$, as a function of $\vctr{y}$, is $L_{i\cm\vctr{x}}$-Lipschitz with respect to the Euclidean metric on $\spcsup{R}{d_{i-1}}$. Let $\sup_{\vctr{x} \in \spcsup{R}{d_{i-1}}} L_{i\cm\vctr{x}} = L_{i}$, which we assume to be finite. 

\begin{proposition}
\label{gaussian complexity kmlp}
Given an $l$-layer kMLP, approximate $\vctrsub{w}{f_i^j}$ using $$\sum_{\nu=1}^m \alpha_{i\cm\nu}^{j} \vfnwasub{\phi}{\vfnsub{F}{i-1}\circ \cdots \circ \vfnwasub{F}{\vctrsub{x}{\nu}}{1}}{i},$$ where the $\vctrsub{x}{\nu}$ are an $m$-subset of $S_{\rvctr{X}}$, $1\leq m\leq N$, $\vctrdub{\boldalpha}{i}{j} \coloneqq \paren{\alpha^{j}_{i\cm 1}\comma \ldots\comma \alpha^{j}_{i\cm m}} \in \spcsup{R}{m}$ and $b_{f_i^{j}}\in \spc{R}$. Assume $\normsub{\vctrdub{\boldalpha}{i}{j}}{1} \leq A_i$ and let $d_l = 1$. Consider $$\spcsub{F}{1} = \setwvert{\vctr{x}\mapsto\paren{\fnwadub{f}{\vctr{x}}{1}{1}\comma \ldots\comma \fnwadub{f}{\vctr{x}}{1}{d_1}}}{\fndub{f}{1}{j} \in \Omega\comma j=1\comma\ldots\comma d_1},$$ where $\Omega$ is a given hypothesis class of functions from $\spcsup{R}{d_0}$ to $\spc{R}$. Denote the class of functions implemented by this kMLP as $\spcsub{F}{l\text{-kMLP}}$, if $\vfnsub{F}{1} \in \spcsub{F}{1}$, for $i\geq 2$, we have
\begin{equation*}
\gcpxwasub{\spcsub{F}{l\text{-kMLP}}}{N} \leq 2d_1 \prod_{i=2}^l A_i L_i d_i \gcpxwasub{\Omega}{N}.
\end{equation*}
\end{proposition}

It is worth noting that the model complexity kMLP grows in the depth and width of the network in a similar way as that of an MLP \citep{Sun2016}. In particular, the expressive power of the model increases linearly in the width of a given layer and roughly exponentially in the depth of the network.

\section{A Layer-Wise Learning Framework} 
\label{learning algorithm}
We now formally present our greedy framework for learning compositional hypothesis classes in a supervised setting. To simplify discussion, we first consider the two-layer case, i.e., 
$$
  \spc{F} = \setwvert{
  \vfn{F} = \vfnsub{F}{2}\circ\vfnsub{F}{1}
  }{
  \vfnsub{F}{i}\in\spcsub{F}{i}\cm i = 1\cm 2
}.
$$
Define $\vfndub{F}{1}{\star}\circ\vfndub{F}{2}{\star}=\vfnsup{F}{\star}\coloneqq\argmin_{\vfn{F}\in\spc{F}}\fnwa{
  \tilde{R}
  }{
  \vfnwa{F}{S_{\vctr{X}}}\cm S_Y
}$. The goal is to learn the input layer to find $\vfndub{F}{1}{\star}$ (without touching the output layer), freeze the input layer afterwards, and then learn the output layer to find $\vfndub{F}{2}{\star}$.

To disentangle the learnings of the two layers, we must disentangle the definitions of $\vfndub{F}{1}{\star}$ and $\vfndub{F}{2}{\star}$. The idea is to re-characterize $\vfndub{F}{1}{\star}$, i.e., to derive conditions under which $\vfnsub{F}{1} = \vfndub{F}{1}{\star}$, using no information on the trainable parameters of the output layer. Then, we need to translate these conditions into choosing a new loss $\ell_1$ (inducing a new risk $R_1$), a function $s$, and a function $u$ accordingly with the property that$$
  \argmin_{\vfnsub{F}{1}\in\spcsub{F}{1}}\fnwasub{
  {R}
  }{
  \fnwa{s}{\vfnwasub{F}{\vctr{X}}{1}}\cm \fnwa{u}{Y}
  }{
  1
}= \vfndub{F}{1}{\star}
$$
and that $s\cm u$ do not rely on the trainable parameters of the output layer. An objective $\fnwasub{
  {\tilde{R}}
  }{
  \fnwa{s}{\vfnwasub{F}{S_\vctr{X}}{1}}\cm \fnwa{u}{S_Y}
  }{
  1
}$ can be subsequently chosen, and we can find $\vfndub{F}{1}{\star}$ by training the input layer to minimize this new objective. This training process requires no tuning on the output layer as this new objective does not involve the trainable parameters of it. 

The re-characterization of $\vfndub{F}{1}{\star}$ is dependent on $\vfnsub{F}{2}$ and $\tilde{R}$. Therefore, different choices induce different realizations of our general framework.

The search of $u$ can be understood as the procedure of explicitly backpropagating the targets $S_Y$ to the hidden layers. This contrasts how learning is made possible in BP via backpropagating derivative information but not the targets directly. 

We proceed by first describing the general framework and then, as examples, provide realizations under a specific choice of $\spcsub{F}{2}$ and two families of objectives. Finally, based on these realizations, we provide a sample layer-wise training algorithm for learning an $l$-layer feedforward network for classification, where $l\geq 2$ can be arbitrary. This layer-wise algorithm is simple to implement and its learning dynamics enjoy an intuitive geometric interpretation. 

\subsection{The Framework}

Let the architecture $\spc{F}$ and objective $\tilde{R}$ be given, our greedy learning framework for the two-layer compositional hypothesis class consists of the following steps:

\begin{enumerate}
  \item Finding $\vfndub{F}{1}{\star}$
    \begin{enumerate}
      \item\label{1a} Define an equivalence relation between hypotheses.
      \item\label{1b} Give an equivalent definition for $\vfndub{F}{1}{\star}$ under the new equivalence relation.
      \item\label{1c} Re-characterize $\vfndub{F}{1}{\star}$ for the given $\spc{F}$ under the given objective $\tilde{R}$.
      \item\label{1d} Choose $s\cm u\cm \ell_1\cm$ and $\fnsub{\tilde{R}}{1}$ accordingly.
      \item Train the input layer to minimize $\fnwasub{\tilde{R}}{\fnwa{s}{\vfnwasub{F}{S_\vctr{X}}{1}}\cm \fnwa{u}{S_Y}}{1}$.
      \item After training, freeze the input layer at, say, $\vfndub{F}{1}{\circ}$.
    \end{enumerate}
  \item Finding $\vfndub{F}{2}{\star}$
    \begin{enumerate}
      \item Train the output layer to minimize $\fnwa{\tilde{R}}{\vfnsub{F}{2}\circ\vfnwadub{
            F
            }{
            S_\vctr{X}
            }{
          1
        }{
        \circ
    }\cm S_Y}$.
    \end{enumerate}
\end{enumerate}

We now provide more details for a couple of the listed steps.

\bigskip

\noindent\textbf{Step~\ref{1a}. Define an equivalence relation between hypotheses}

In our framework, we use the following definition of equivalence between hypotheses of the input layer: $$
\vfnsub{F}{1} = \vfnsub{G}{1} \quad\text{ if and only if }\quad \min_{\vfnsub{F}{2}\in\spcsub{F}{2}} \fnwa{
  \tilde{R}
  }{
  \vfnsub{F}{2}\circ\vfnsub{F}{1}
}=\min_{\vfnsub{F}{2}\in\spcsub{F}{2}} \fnwa{
  \tilde{R}
  }{
  \vfnsub{F}{2}\circ\vfnsub{G}{1}
}\cm\forall S.
$$
It is easy to check that this is indeed an equivalence relation. Intuitively, this means that we consider two hypotheses of the input layer to be equally good if the best networks one can build with these two hypotheses minimize the objective function equally well, i.e., when they have the same ``potential''. Evidently, this notion of equivalence is proper and sufficient as we have no knowledge of $\vfnsub{F}{2}$ while we train the input layer. 

\bigskip

\noindent\textbf{Step~\ref{1b}. Give an equivalent definition for $\vfndub{F}{1}{\star}$ under the new equivalence relation}

Compared to the original minimizer definition of $\vfndub{F}{1}{\star}$, it is easier to work with the following more concrete definition under the equivalence relation described in Step~\ref{1a}.
\begin{lemma} 
\label{equivalent def of layer-wise opt}
Suppose $\vfndub{F}{1}{\star} \in \spcdub{F}{1}{\prime} \subseteq \spcsub{F}{1}$ and $\vfndub{F}{2}{\star} \in \spcdub{F}{2}{\prime} \subseteq \spcsub{F}{2}$, we have
\begin{equation*}
\vfndub{F}{1}{\star} = \argmin_{\vfnsub{F}{1} \in \spcdub{F}{1}{\prime}} \min_{\vfnsub{F}{2} \in \spcdub{F}{2}{\prime}} \fnwa{\tilde{R}}{\vfnsub{F}{2} \circ \vfnsub{F}{1}}.
\end{equation*}
\end{lemma}
This definition is easier to work with when we later re-characterize $\vfndub{F}{1}{\star}$ because it shrinks the range of $\vfnsub{F}{2}$ we need to consider for each $\vfnsub{F}{1}$ to only the minimizer $\vfnsub{F}{2}$ under that specific $\vfnsub{F}{1}$.

\subsection{Some Realizations}

We now provide realizations of the greedy learning framework under a specific family of $\spcsub{F}{2}$ and two classes of objective functions. Note that these realizations are certainly not all that can be derived from our layer-wise framework. We leave the exploration of more such realizations as future work.

Steps~\ref{1a} and \ref{1b} are the same for all realizations. Therefore, the only nontrivial steps in our framework to discuss for specific realizations are steps~\ref{1c} and \ref{1d}. 

The specific $\spcsub{F}{2}$ we consider in this section is defined as the set of functions of the following form:
$
    \vfnwasub{
        F
        }{
        \vctr{x}
        }{
        2
    } = 
(f_2^{1}(\vctr{x})\cm\ldots\cm f_2^{d_2}(\vctr{x}))$,
$f_2^{j}(\vctr{x}) = 
    \innersub{
        \vctrsub{w}{f_2^{j}}
        }{
    \phi(\vctr{x})
        }{
        H
    } + b_{f_2^{j}}$ with kernel $\kernelwa{
    \vctr{x}
    }{
    \vctr{y}
    } = \innersub{
    \fnwa{\phi}{\vctr{x}}
    }{
    \fnwa{\phi}{\vctr{y}}
    }{
    H
  }$, $j=1\cm\ldots\cm d_2\cm\vctr{x}\cm\vctr{y}\in\spcsup{R}{d_0}$, where we have omitted and will continue to omit writing out explicitly the function composition: for example, for $\vctr{x}\in\spcsup{R}{d_0}$, we write $
    \vfnwasub{
        F
        }{
        \vctr{x}
        }{
        2
    }
    $ in place of $
    \vfnwasub{
        F
        }{\vfnwasub{
            F
        }{
        \vctr{x}
        }{
        1
    }
        }{
        2
    }
    $. There is no assumption needed on $\spcsub{F}{1}$.

    For these realizations, we consider the case $Y\in\setonly{+1\cm -1}$, and we shall use subscript $+$ or $-$ to indicate the class of a particular example, if needed.

\bigskip

\noindent\textbf{Re-characterize $\vfndub{F}{1}{\star}$ under regularized hinge loss as objective}
\label{optimal l-1} 

Let $d_2 = 1$, write $\fnsub{f}{2}$ in place of $\vfnsub{F}{2}$ accordingly. Let the objective function $\fnwa{\tilde{R}}{\fnsub{f}{2}\circ\vfnsub{F}{1}}$ be $\fnwa{\hat{R}}{\fnsub{f}{2}\circ\vfnsub{F}{1}} + \tau \normsub{\vctrsub{w}{\fnsub{f}{2}}}{H}$, where $\tau> 0$ is a hyperparameter that can be chosen as desired and $$\fnwa{\hat{R}}{\fnsub{f}{2}\circ\vfnsub{F}{1}} = \frac{1}{N}\sum_{n=1}^N \fnwa{\ell}{\fnsub{f}{2}\circ\vfnsub{F}{1}\comma \paren{\vctrsub{x}{n}\comma y_n}}$$ with $\fnwa{\ell}{\fnsub{f}{2}\circ\vfnsub{F}{1}\comma \paren{\vctrsub{x}{n}\comma y_n}} = \max\paren{0, 1 - y_n \fnwasub{f}{\vctrsub{x}{n}}{2}}$, the hinge loss. Let $\kappa = \frac{1}{N}\sum_{n=1}^N \ind{y_n = +}$. We now re-characterize $\vfndub{F}{1}{\star}$.
\begin{theorem} 
\label{lemma6}
Assume that $\tau < \sqrt{2(c - a)}\min\paren{\kappa\comma 1 - \kappa}$ and that there exist $\paren{\vctrsub{x}{+}\comma y_+}\comma \paren{\vctrsub{x}{-}\comma y_-} \in S$ such that $\fnwa{\ell}{\fndub{f}{2}{\star}\circ\vfndub{F}{1}{\star}\comma \paren{\vctrsub{x}{n}\comma y_n}} = 0$, $n = +\comma -$.

If $\vfnsub{F}{1}$ satisfies
\begin{equation}
\label{eq1}
\begin{split}
&\kernelwa{\vfnwasub{F}{\vctrsub{x}{+}}{1}}{\vfnwasub{F}{\vctrsub{x}{-}}{1}} = a \quad \text{and} \quad \\
&\kernelwa{\vfnwasub{F}{\vctr{x}}{1}}{\vfnwasub{F}{\vctrsup{x}{\prime}}{1}} = c
\end{split}
\end{equation}
for all pairs of $\vctrsub{x}{+}\comma \vctrsub{x}{-} \in S_{\rvctr{X}}$ and all pairs of $\vctr{x}\comma \vctrsup{x}{\prime}\in S_\vctr{X}$ with $y = y'$, then $\vfnsub{F}{1} = \vfndub{F}{1}{\star}$.
\end{theorem}

\bigskip

\noindent\textbf{Re-characterize $\vfndub{F}{1}{\star}$ under regularized supervised representation similarity (SRS) loss as objective}

Consider function $h:\spcsup{R}{d_2}\times\spcsup{R}{d_2}\to\spc{R}$ with the property that $\fnwa{h}{\vctr{x}\cm \vctr{y}}$, as a function of $\vctr{x}$ and $\vctr{y}$, has the following properties:
\begin{itemize}
  \item $\inf_{\vctr{x}\cm\vctr{y}} \fnwa{h}{\vctr{x}\cm\vctr{y}} = b>-\infty$, $\sup_{\vctr{x}\cm\vctr{y}} \fnwa{h}{\vctr{x}\cm\vctr{y}} = d<\infty$, $d>b$;
    \item $h$ depends only on $
        \normsub{
            \vctr{x} - \vctr{y}
            }{
            q
          }$ for some $q\geq 1$, i.e., $\fnwa{h}{\vctr{x}\cm\vctr{y}} = \fnwa{h}{\normsub{\vctr{x} - \vctr{y}}{q}}$; 
    \item $h$ strictly decreases in $
        \normsub{
            \vctr{x} - \vctr{y}
            }{
            q
        }$ for all $\vctr{x}\cm \vctr{y}\in 
        \spcsup{
            R
            }{
            d_2
          }\text{ with } 
          \fnwa{h}{
            \vctr{x}
            \cm
            \vctr{y}
            }>b$.
\end{itemize}

Define the following SRS loss:
\begin{align*}
\fnwa{
    \ell
    }{
    \vfnsub{F}{2}\circ\vfnsub{F}{1}\cm
    (\vctr{x}\cm y)\cm
    (\vctrsup{x}{\prime}\cm y^\prime)
    }= 
\absolute{
    \fnwa{
        g
        }{
        y\cm y^\prime  
        } -
        \fnwa{h}{
        \vfnwasub{
            F
            }{
            \vctr{x}
            }{
            2
        }\cm{
        \vfnwasub{
            F
            }{
            \vctrsup{x}{\prime}
            }{
            2
        }
        }
        }}^p,
\end{align*}
where $p\geq 1$ can be arbitrarily chosen and $
\fnwa{
        g
        }{
        y\cm y^\prime  
        } = b$ if $y \neq y^\prime$ and $d$ if otherwise. 

It is easy to see that this loss penalizes the similarity between images of examples under the mapping $\vfnsub{F}{2}\circ\vfnsub{F}{1}$ based on their classes, therefore the name supervised representation similarity.

Let the objective function be
\begin{align*}
&\fnwa{
    \tilde{R}
    }{
    \vfnsub{F}{2}\circ\vfnsub{F}{1}
    }\\ &\quad= 
\frac{1}{N^2}
\sum_{n\cm m=1}^N
\fnwa{
    \ell
    }{
    \vfnsub{F}{2}\circ\vfnsub{F}{1}\cm
    (\vctrsub{x}{n}\cm y_n)\cm
    (\vctrsub{x}{m}\cm y_m)
    }+ 
\tau
\fnwa{
    t
    }{
    \normsub{\vctrsub{w}{f_2^{1}}}{H}\cm\ldots\cm
    \normsub{\vctrsub{w}{f_2^{d_2}}}{H}
},
\end{align*}
where $\tau>0$ can be freely chosen and $t$ can be any function that strictly decreases in all of its arguments.

\begin{theorem}
    \label{new theorem 4.5}
    Assume that there exist $
    (\vctrsub{x}{+}\cm y_+)\cm
    (\vctrsub{x}{-}\cm y_-)\in S$ such that
    
    $\fnwa{
        \ell
        }{
      \vfndub{F}{2}{\star}\circ\vfndub{F}{1}{\star}\cm
        (\vctrsub{x}{+}\cm y_+)\cm
        (\vctrsub{x}{-}\cm y_-)
        } = 0$.
    Also assume that for all $j$, $
    \normsub{
        \vctrsub{w}{f_2^{j\star}}
        }{
        H
    }>0$.

    If $\vfnsub{F}{1}$ satisfies
    \begin{equation}
        \begin{split}
        \label{eq100} 
        &\kernelwa{
            \vfnwasub{
                F
                }{
                \vctrsub{x}{+}
                }{
                1
            }
            }{
            \vfnwasub{
                F
                }{
                \vctrsub{x}{-}
                }{
                1
            }
            }
             = a; \text{ and }\\
        &\kernelwa{
            \vfnwasub{
                F
                }{
                \vctr{x}
                }{
                1
            }
            }{
            \vfnwasub{
                F
                }{
                \vctrsup{x}{\prime}
                }{
                1
            }
            }= c
        \end{split}
    \end{equation}
for all pairs of $\vctrsub{x}{+}\cm \vctrsub{x}{-}\in S_{\rvctr{X}}$ and all pairs of $\vctr{x}\cm \vctrsup{x}{\prime}\in S_{\rvctr{X}}$ with $y=y^\prime$, then $\vfnsub{F}{1} = \vfndub{F}{1}{\star}$.
\end{theorem}

\bigskip

\noindent\textbf{On selecting $s\cm u\cm\ell_1$, and $\tilde{R}_1$}

For both of the two objectives described above, we may choose, for example, $s$ to be the kernel function $k$, $u$ to be the function defined as $
\fnwa{
        u
        }{
        y\cm y^\prime  
      } = a$ if $y \neq y^\prime$ and $c$ if otherwise, and $\ell_1$ to be the SRS loss defined earlier with $g$ set to $u$ and $h$ set to $s$, i.e., 
\begin{align*}
\fnwasub{
    \ell
    }{
    \vfnsub{F}{1}\cm
    (\vctr{x}\cm y)\cm
    (\vctrsup{x}{\prime}\cm y^\prime)
  }{
  1
}= 
\absolute{
    \fnwa{
        u
        }{
        y\cm y^\prime  
        } - 
        \kernelwa{
        \vfnwasub{
            F
            }{
            \vctr{x}
            }{
            1 
        }}{
        \vfnwasub{
            F
            }{
            \vctrsup{x}{\prime}
            }{
            1
        }
        }
}^p,
\end{align*}
where $p\geq 1$ can be freely chosen. This of course requires $k$ to satisfy the aforementioned conditions on $h$ for the resulting loss to be a valid SRS loss. 

Under this selection of $\ell_1$, it is evident that the minimizers of $\ell_1$ (and also $R_1$) are all equal to $\vfndub{F}{1}{\star}$ by Theorems~\ref{lemma6} and \ref{new theorem 4.5}. $\tilde{R}_1$ can be set to the empirical SRS loss plus an arbitrary regularization term on norms of the weights.

\bigskip

\noindent\textbf{Generalizing to $l$-layer feedforward models with $l\geq 2$}

The generalization to a feedforward model with $l$ layers, where $l\geq 2$ can be arbitrary, is trivial. To begin with, treat $\vfnsub{F}{l}$ and $\vfnsub{F}{l-1}\circ\cdots\circ\vfnsub{F}{1}$ as the earlier $\vfnsub{F}{2}$ and $\vfnsub{F}{1}$, respectively. Then work as in the two-layer case to find an objective $\fnsub{\tilde{R}}{l-1}$ for $\vfnsub{F}{l-1}\circ\cdots\circ\vfnsub{F}{1}$. This reduces the $l$-layer problem to an $l-1$-layer problem. Repeat this procedure on the rest of the layers until we return to the original two-layer case.

\subsection{A Layer-Wise Training Algorithm for an $l$-Layer ($l\geq 2$) Feedforward Network for Classification}
\label{sample algorithm}

We can build upon the above realizations an certified (in the sense that the optimality is guaranteed) layer-wise algorithm for training an $l$-layer ($l\geq 2$) feedforward network for classification tasks. In this section, we describe this algorithm and show that it enjoys a geometric interpretation that makes the learning dynamics transparent. Moreover, we show that there is a simple acceleration method for the kernelized non-input layers, making the architecture more practical.

We present this algorithm for binary classification. Nevertheless, as multi-class problems can be reduced to a set of binary classification problems by using either the one-vs-all or the one-vs-one strategy \citep{Scholkopf01}, an extension of this algorithm to multi-class problems is trivial. 

The architecture considered is as follows:
\begin{align*}
  &\Bigg\{
  f_l\circ\cdots\circ\vfnsub{F}{1}: \spcsup{R}{d_0}\to\spc{R}
    \mid\\
    &\qquad\vfnwasub{
        F
        }{
        \vctr{x}
        }{
        i
    } = 
(f_i^{1}(\vctr{x})\cm\ldots\cm f_i^{d_i}(\vctr{x})),
f_i^{j}(\vctr{x}) = 
    \innersub{
        \vctrsub{w}{f_i^{j}}
        }{
    \phi_i(\vctr{x})
        }{
        H_i
      } + b_{f_i^{j}}\cm\vctrsub{w}{f_i^{j}}\in H_i\cm b_{f_i^j}\in\spc{R}
  \cm\\&\qquad\forall l>i>1\cm\forall j\cm f_l(\vctr{x}) = \innersub{
        \vctrsub{w}{f_l}
        }{
        \phi_l(\vctr{x})
        }{
        H_l
      } + b_{f_l}\cm \vctrsub{w}{f_l}\in H_l\cm b_{f_l}\in\spc{R}
\Bigg\}.
\end{align*}

Note that we have made no assumption on the input layer.

For $i<l$, define $$
\fnwasub{\tilde{R}}{\vfnsub{F}{i}}{i} = \frac{1}{N^2}\sum_{n\cm m}^N
\fnwasub{
    \ell
    }{
    \vfnsub{F}{i}\cm
    (\vctrsub{x}{n}\cm y_n)\cm
    (\vctrsub{x}{m}\cm y_m)
  }{
  i
} + 
\tau_i
\fnwa{
    t
    }{
    \normsub{\vctrsub{w}{f_i^{1}}}{H_i}\cm\ldots\cm
    \normsub{\vctrsub{w}{f_i^{d_i}}}{H_i}
},
$$
where
\begin{align*}
\fnwasub{
    \ell
    }{
    \vfnsub{F}{i}\cm
    (\vctrsub{x}{n}\cm y_n)\cm
    (\vctrsub{x}{m}\cm y_m)
  }{
  i
}= 
\absolute{
    \fnwa{
        u
        }{
        y_n\cm y_m 
        } - 
    \kernelwasub{
        \vfnwasub{
            F
            }{
            \vctrsub{x}{n}
            }{
            i 
        }
        }{
        \vfnwasub{
            F
            }{
            \vctrsub{x}{m}
            }{
            i
        }
        }{
        i+1
    }
}^p,
\end{align*}
$p\geq 1$ can be chosen freely, $\tau>0$, $\fnwa{u}{y\cm y^\prime} = a$ if $y = y^\prime$ and $c$ otherwise, and $t$ can be any function that strictly decreases in all of its arguments.

For $i=l$, define :$$
\fnwasub{\tilde{R}}{\fnsub{f}{l}}{l}=
\frac{1}{N}\sum_{n=1}^N \fnwasub{\ell}{\fnsub{f}{l}\comma \paren{\vctrsub{x}{n}\comma y_n}}{l} + \tau_l \normsub{\vctrsub{w}{\fnsub{f}{l}}}{H_l},$$ where $$
\fnwasub{\ell}{\fnsub{f}{l}\comma \paren{\vctrsub{x}{n}\comma y_n}}{l} = \max\paren{0, 1 - y_n \fnwasub{f}{\vctrsub{x}{n}}{l}}
$$
Then the training algorithm is given in Algorithm~\ref{alg1}.

\begin{algorithm}
\caption{A certified layer-wise training algorithm for classification.}
\label{alg1}
\begin{algorithmic}
\STATE \textbf{input:} training set $\{(\vctr{x}_n, y_n)\}_{n=1}^N$
\STATE \textbf{initialize:} initialize and freeze all layers 
\FOR{$i = 1\cm 2\cm \ldots\cm l$}
\STATE unfreeze layer $i$
\STATE train layer $i$ to minimize $\fnsub{\tilde{R}}{i}$
\STATE freeze layer $i$
\ENDFOR
\end{algorithmic}
\end{algorithm}

The optimality of this training algorithm is justified by Theorems~\ref{lemma6} and \ref{new theorem 4.5} when the $\tau_i$ and $k_i$ satisfy the corresponding conditions for all $i=1\cm\ldots\cm l$.

We emphasize that this particular training algorithm gives great freedom to the choice of $\vfnsub{F}{1}$: it can be any arbitrary architecture. In particular, it can be a stack of multiple layers in practice. This stack can be trained with an end-to-end method such as BP.

\subsubsection{Geometric Interpretation of Learning Dynamics}
\label{learning dynamics}
The sufficient conditions described by Eq.~\ref{eq1} and Eq.~\ref{eq100} can be interpreted geometrically: under an $\vfnsub{F}{1}$ satisfying these conditions, images of examples from distinct classes are as distant as possible in the RKHS induced by $k$ whereas images of examples from the same class are as concentrated as possible (see proof of Theorem~\ref{lemma6} in Appendix~\ref{proofs}). Intuitively, such a representation is the ``easiest'' for the classification task. And our earlier theorems essentially justified this intuition in a rigorous fashion. 

Therefore, the learning dynamics of this training algorithm can be given a straightforward geometric interpretation: it trains each layer to push apart examples from different classes while squeeze together those within the same class. In other words, each layer learns a better representation of the data. Eventually, the output layer works as a classifier on the final hidden representation.

\subsubsection{Accelerating the Kernelized Layers}
There is a natural method to accelerate the kernelized non-input layers: the hidden targets are sparse in the sense that for $1\leq i<l$ and any $\vfnsub{F}{i}$ satisfying Eq.~\ref{eq1} or Eq.~\ref{eq100}, we have $\vfnwasub{\phi}{\vfnwasub{F}{\vctrsub{x}{m}}{i}}{i+1} = \vfnwasub{\phi}{\vfnwasub{F}{\vctrsub{x}{n}}{i}}{i+1}$ if $y_m = y_n$ and $\vfnwasub{\phi}{\vfnwasub{F}{\vctrsub{x}{m}}{i}}{i+1} \neq \vfnwasub{\phi}{\vfnwasub{F}{\vctrsub{x}{n}}{i}}{i+1}$ if $y_m \neq y_n$ (see proof of Theorem~\ref{lemma6} in Appendix~\ref{proofs}). Since we usually approximate $\vctrdub{w}{i+1}{j}$ using $\sum_{n=1}^N \alpha_{i+1\cm n}^{j} \vfnwasub{\phi}{\vfnwasub{F}{\vctrsub{x}{n}}{i}}{i+1}$, retaining only one example from each class would result in exactly the same hypothesis class $\spcsub{F}{i+1}$ because $\setwvert{\sum_{n=1}^N \alpha_{i+1\cm n}^{j} \vfnwasub{\phi}{\vfnwasub{F}{\vctrsub{x}{n}}{i}}{i+1}}{\alpha_{i+1\cm n}^{j} \in \spc{R}} = \setwvert{\sum_{n=+\comma -} \alpha_{i+1\cm n}^{j} \vfnwasub{\phi}{\vfnwasub{F}{\vctrsub{x}{n}}{i}}{i+1}}{\alpha_{i+1\cm n}^{j} \in \spc{R}}$ for arbitrary $\vctrsub{x}{+}\comma \vctrsub{x}{-}$ in $S_{\rvctr{X}}$. 

Thus, after training a given layer, depending on how well its objective function has been minimized, one may discard some of the centers for kernel machines of the next layer to speed up the training of that layer without sacrificing performance. This trick also has a regularization effect on the kernel machines since the number of trainable parameters of a kernel machine grows linearly in the number of its centers. 

\subsection{How is our layer-wise framework different from the existing layer-wise pre-training schemes?}

Existing layer-wise pre-training methods such as those proposed in \citep{Hinton06b} and \citep{Bengio07} require backpropagation (BP) fine-tuning. This is because, to the best of our knowledge, no optimality guarantee comparable to that provided by BP can be made for these pre-training algorithms. In other words, the layer-wise pre-training commonly used in the deep learning community does not necessarily learn the hypothesis that minimizes the objective function for the network and thus can only be used as an add-on to BP that helps BP converge faster.

In contrast, our work proves such optimality for our layer-wise training scheme in certain specific learning settings and therefore completely removes the need for BP in these settings. To put this in another way, even if one applies BP after performing our layer-wise training, one will not (in theory) end up with a hypothesis that is strictly better than the one learned by the layer-wise learning process in terms of minimizing the objective function of the network.

Coming up with a purely layer-wise substitute for BP is relevant because, as we have mentioned, BP can be computationally expensive and its end-to-end nature makes it practically impossible to precisely trace the source of unsatisfying performance and find out which layer or layers is to be blamed. This can make the architecture search process lengthy and sometimes painful. Furthermore, training all layers simultaneously complicates the parameter space and may introduce more local minima to the learning process, which can be another unwanted factor for gradient descent-based learning. In contrast, a fully layer-wise training process allows one to divide and conquer the learning problem and reveals more useful information about training, mitigating the aforementioned issues to some extent. 

\section{Related Works}
The link between NNs and the kernel method has been long known. In \citep{Vapnik2000}, the hyperbolic tangent kernel was defined and used in SVM, leading to an architecture equivalent to a shallow MLP. \cite{suykens1999training} viewed MLP as SVM by treating the hidden layer as the feature map and proposed accordingly a modified support vector method to train the former. More recently, \cite{Cho09} defined an ``arc cosine'' kernel to imitate the computations performed by a one-layer MLP. \cite{Zhuang11} extended the idea to arbitrary kernels with a focus on MKL, using an architecture similar to a two-layer kMLP. As a further generalization, \cite{Zhang2017} proposed kMLP and fully-kernelized CNN. However, they did not extend the idea to more network architectures. These works essentially combine kernel method with deep learning by substituting neurons in NNs with kernel machines, which is similar to what we are pursuing in this work. However, to the best of our knowledge, our work enjoys perhaps the greatest generality among works that follow this line of research. 

There are also works that attempt to integrate kernel method with deep learning using other methods. \cite{suykens2017deep} drew connections between restricted Boltzmann machines (RBM) and kernel machines by creating RBM-like representations for the latter. The resulting restricted kernel machines (RKMs) are then combined to form deep RKMs. \cite{Mairal2014} proposed to learn hierarchical representations by learning mappings of kernels that are invariant to irrelevant variations in images. \cite{hermans2012recurrent} used the kernel method to expand the echo state networks to essentially infinite-sized recurrent neural networks. The resulting network can then be viewed as a recursive kernel that can be used in SVMs. \cite{Wilson2016} proposed to learn the covariance matrix of a Gaussian process using an NN in order to make the kernel ``adaptive''. Such an interpretation of ``adaptive'' kernels can be given to KNs as well. This idea also underlies the now standard approach of combining a deep NN with SVM for classification, which was first explored by \cite{Huang2006} and \cite{Tang2013} and can be viewed as a special case of the proposed kernelization framework. In terms of the training of such hybrid systems, there are mainly two methods. The first is to apply BP to the entire model \citep{Tang2013}, which enjoys an optimality guarantee from BP but forces the SVM to be trained with gradient descent instead of the more efficient optimization algorithms that are usually used for SVMs. The alternative is to feed the hidden representations from a trained NN to the SVM and train the latter in the usual way \citep{Huang2006}, but this practice is not theoretically solid. The proposed layer-wise learning framework serves as another alternative that combines the best of both worlds: one can train the NN and SVM separately with an optimality guarantee as that given by BP.

Much works have been done to improve or substitute BP in learning a deep architecture. Most aim at improving the classical method, working as add-ons for BP. The most notable ones are perhaps the unsupervised greedy pre-training techniques proposed by \cite{Hinton06b} and \cite{Bengio07}. Among works that try to completely substitute BP, none provided a comparable optimality guarantee in theory as that given by BP. \cite{Fahlman1990} pioneered the idea of greedily learn the architecture of an NN. In their work, each new node is added to maximize the correlation between its output and the residual error signal. Several authors explored the idea of approximating error signals propagated by BP locally at each layer or each node \citep{Bengio2014, Carreira2014, Lee2015, Balduzzi2015, Jaderberg2016}. \cite{Zhou2017} proposed a BP-free deep architecture based on decision trees. \cite{Raghu2017} attempted to quantify the quality of hidden representations toward learning more interpretable deep architectures, sharing a motivation similar to ours. 

\section{Experiments} 
\label{experiments}

We now demonstrate the competence of the kernelized models and the effectiveness of the proposed layer-wise framework via experiments. We will be implementing the sample training algorithm described in Section~\ref{sample algorithm} throughout. This section will be divided into two parts. The first one will be dedicated to comparing KNs with traditional kernel machines. In the second part, we compare KNs with other popular connectionist models. These empirical results serve as proofs of concept for the proposed architectures as well as the greedy training framework. 

\subsection{Comparing KNs with Classical Kernel Machines} 
\label{kn vs mkl}
We now compare a single-hidden-layer kMLP using simple, generic kernels with the classical SVM and SVMs enhanced by MKL algorithms that used significantly more kernels to demonstrate the competence of kMLP and in particular, its ability to perform well without excessive kernel parameterization. The standard SVM and seven other SVMs enhanced by popular MKL methods were compared~\citep{Zhuang11}, including the classical convex MKL~\citep{Lanckriet04} with kernels learned using the extended level method proposed in~\citep{Xu09} ($\text{MKL}^{\text{LEVEL}}$); MKL with $L^p$ norm regularization over kernel weights~\citep{Kloft11} ($L^p$MKL), for which the cutting plane algorithm with second order Taylor approximation of $L^p$ was adopted; Generalized MKL in~\citep{Varma09} (GMKL), for which the target kernel class was the Hadamard product of single Gaussian kernel defined on each dimension; Infinite Kernel Learning in~\citep{Gehler08} (IKL) with $\text{MKL}^{\text{LEVEL}}$ as the embedded optimizer for kernel weights; 2-layer Multilayer Kernel Machine in~\citep{Cho09} (MKM); 2-Layer MKL (2LMKL) and Infinite 2-Layer MKL in~\citep{Zhuang11} ($\text{2LMKL}^{\text{INF}}$).

Eleven binary classification data sets that have been widely used in MKL literature were split evenly for training and test and were all normalized to zero mean and unit variance prior to training. Twenty runs with identical settings but random weight initializations were repeated for each model. For each repetition, a new training-test split was selected randomly.

For kMLP, all results were achieved using a greedily-trained, one-hidden-layer model with the number of kernel machines ranging from 3 to 10 on the first layer for different data sets. The second layer was a single kernel machine. All kernel machines within one layer used the same Gaussian kernel ($\kernelwa{\vctr{x}}{\vctr{y}} = e^{-\normsub{\vctr{x} - \vctr{y}}{2}/\sigma^2}$), and the two kernels on the two layers differed only in kernel width $\sigma$. All hyperparameters were chosen via $5$-fold cross-validation. As for the other models compared, for each data set, SVM used a Gaussian kernel. For the MKL algorithms, the base kernels contained Gaussian kernels with 10 different widths on all features and on each single feature and polynomial kernels of degree 1 to 3 on all features and on each single feature. For $\text{2LMKL}^{\text{INF}}$, one Gaussian kernel was added to the base kernels at each iteration. Each base kernel matrix was normalized to unit trace. For $L^p$MKL, $p$ was selected from $\{2, 3, 4\}$. For MKM, the degree parameter was chosen from $\{0, 1, 2\}$. All hyperparameters were selected via $5$-fold cross-validation.
\begin{table}
\caption{Average test error ($\%$) and standard deviation ($\%$) from $20$ runs. Results with overlapping $95\%$ confidence intervals (not shown) are considered equally good. Best results are marked in bold. The average ranks (calculated using average test error) are provided in the bottom row. When computing confidence intervals, due to the limited sizes of the data sets, we pooled the twenty random samples.}
\label{table1}
\begin{center}
\begin{adjustbox}{width=1\textwidth}
\begin{tabular}{lccccccccccccr} 
\toprule
 & Size/Dimension & SVM & $\text{MKL}^{\text{LEVEL}}$ & $L^p$MKL & GMKL & IKL & MKM & 2LMKL & $\text{2LMKL}^{\text{INF}}$ & $\text{kMLP-1}$\\
\midrule
Breast & 683/10& 3.2$\pm$ 1.0& 3.5$\pm$ 0.8& 3.8$\pm$ 0.7& 3.0$\pm$ 1.0& 3.5$\pm$ 0.7& 2.9$\pm$ 1.0& 3.0$\pm$ 1.0& 3.1$\pm$ 0.7& \textbf{2.4$\pm$ 0.7}\\ 
 
Diabetes & 768/8& \textbf{23.3$\pm$ 1.8}& 24.2$\pm$ 2.5& 27.4$\pm$ 2.5& 33.6$\pm$ 2.5& \textbf{24.0$\pm$ 3.0}& 24.2$\pm$ 2.5& \textbf{23.4$\pm$ 1.6}& \textbf{23.4$\pm$ 1.9}& \textbf{23.2$\pm$ 1.9}\\
 
Australian & 690/14& 15.4$\pm$ 1.4& 15.0$\pm$ 1.5& 15.5$\pm$ 1.6& 20.0$\pm$ 2.3& \textbf{14.6$\pm$ 1.2}& 14.7$\pm$ 0.9& \textbf{14.5$\pm$ 1.6}& \textbf{14.3$\pm$ 1.6}& \textbf{13.8$\pm$ 1.7}\\ 

Iono & 351/33& 7.2$\pm$ 2.0& 8.3$\pm$ 1.9& 7.4$\pm$ 1.4& 7.3$\pm$ 1.8& 6.3$\pm$ 1.0& 8.3$\pm$ 2.7& 7.7$\pm$ 1.5& \textbf{5.6$\pm$ 0.9}& \textbf{5.0$\pm$ 1.4}\\ 

Ringnorm & 400/20& \textbf{1.5$\pm$ 0.7}& \textbf{1.9$\pm$ 0.8}& 3.3$\pm$ 1.0& 2.5$\pm$ 1.0& \textbf{1.5$\pm$ 0.7}& 2.3$\pm$ 1.0& 2.1$\pm$ 0.8& \textbf{1.5$\pm$ 0.8}& \textbf{1.5$\pm$ 0.6}\\ 

Heart & 270/13& 17.9$\pm$ 3.0& 17.0$\pm$ 2.9& 23.3$\pm$ 3.8& 23.0$\pm$ 3.6& \textbf{16.7$\pm$ 2.1}& 17.6$\pm$ 2.5& \textbf{16.9$\pm$ 2.5}& \textbf{16.4$\pm$ 2.1}& \textbf{15.5$\pm$ 2.7}\\ 

Thyroid & 140/5& 6.1$\pm$ 2.9& 7.1$\pm$ 2.9& 6.9$\pm$ 2.2& 5.4$\pm$ 2.1& 5.2$\pm$ 2.0& 7.4$\pm$ 3.0& 6.6$\pm$ 3.1& 5.2$\pm$ 2.2& \textbf{3.8$\pm$ 2.1}\\ 

Liver & 345/6& \textbf{29.5$\pm$ 4.1}& 37.7$\pm$ 4.5& 30.6$\pm$ 2.9& 36.4$\pm$ 2.6& 40.0$\pm$ 2.9& \textbf{29.9$\pm$ 3.6}& 34.0$\pm$ 3.4& 37.3$\pm$ 3.1& \textbf{28.9$\pm$ 2.9}\\  

German & 1000/24& \textbf{24.8$\pm$ 1.9}& 28.6$\pm$ 2.8& 25.7$\pm$ 1.4& 29.6$\pm$ 1.6& 30.0$\pm$ 1.5& \textbf{24.3$\pm$ 2.3}& 25.2$\pm$ 1.8& 25.8$\pm$ 2.0& \textbf{24.0$\pm$ 1.8}\\  

Waveform & 400/21& 11.0$\pm$ 1.8& 11.8$\pm$ 1.6& 11.1$\pm$ 2.0& 11.8$\pm$ 1.8& \textbf{10.3$\pm$ 2.3}& \textbf{10.0$\pm$ 1.6}& 11.3$\pm$ 1.9& \textbf{9.6$\pm$ 1.6}& \textbf{10.3$\pm$ 1.9}\\

Banana & 400/2& \textbf{10.3$\pm$ 1.5}& \textbf{9.8$\pm$ 2.0}& 12.5$\pm$ 2.6& 16.6$\pm$ 2.7& \textbf{9.8$\pm$ 1.8}& 19.5$\pm$ 5.3& 13.2$\pm$ 2.1& \textbf{9.8$\pm$ 1.6}& 11.5$\pm$ 1.9\\
\midrule
Rank & - & 4.2& 6.3& 7.0& 6.9& 4.3& 5.4& 5.0& 2.8& 1.6\\
\bottomrule
\end{tabular}
\end{adjustbox}
\end{center}
\end{table}

From Table~\ref{table1}, kMLP compares favorably with other models, which validates our claim that kMLP learns its own kernels nonparametrically hence can work well even without excessive kernel parameterization. Performance difference among models can be small for some data sets, which is expected since these datasets are all rather small in size and not too challenging. Nevertheless, it is worth noting that only two Gaussian kernels were used for kMLP, whereas all other models except for SVM used significantly more kernels.  

\subsection{Comparing KNs with NNs}

In this section, we provide empirical results on comparing KN with NN. In the first part, we demonstrate the competence of kernelized NNs and the effectiveness of the layer-wise learning method using kMLPs. We use the proposed layer-wise algorithm derived from our greedy learning framework and Adam~\citep{Kingma2014} as the underlying optimization algorithm. First, we show that this algorithm, albeit only having been certified under certain families of objectives, works well with most popular objective functions in practice. We then compare kMLPs trained with BP and the layer-wise algorithm to show the effectiveness of the latter. Finally, to further showcase the competence of the greedily-trained kernelized models, we compare kMLPs learned layer-wise with other popular deep architectures including MLPs, Deep Belief Networks (DBNs) \citep{Hinton06a} and Stacked Autoencoders (SAEs) \citep{Vincent10}, with the last two trained using a combination of unsupervised greedy pre-training and standard BP \citep{Hinton06b, Bengio07}. We also visualize the learning dynamics of greedy kMLPs and show that it is intuitive and simple to interpret. In the second part of the experiments, we partially kernelize the classic LeNet-5~\citep{Lecun1998} and compare it with the original to validate our claim that the proposed kernelization and training algorithm is flexible in the sense that it works well with any given feedforward NN architecture and one can freely decide the degree of kernelization. The hidden representations learned from the two models are visualized. We show that the hidden representations learned by the kernelized model are much more discriminative than that from the original.

\subsubsection{Part 1: Kernelizing MLPs}

\begin{figure}
\vskip 0.2in
\begin{center}
\centerline{\includegraphics[width=.9\linewidth]{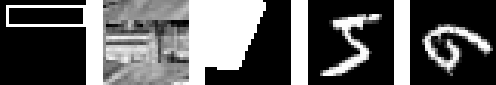}}
\caption{From left to right: example from \emph{rectangles}, \emph{rectangles-image}, \emph{convex}, \emph{mnist (50k test)} and \emph{mnist (50k test) rotated}.}
\label{samples}
\end{center}
\end{figure}

In terms of the datasets used. \emph{rectangles, rectangles-image} and \emph{convex} are binary classification datasets, \emph{mnist (50k test)} and \emph{mnist (50k test) rotated} are variants of MNIST. \emph{fashion-mnist} is the Fashion-MNIST dataset \citep{Xiao2017}. These datasets all contain $28\times28$ grayscale images. In \emph{rectangles, rectangles-image}, the model needs to learn if the height of the rectangle is longer than the width, and in \emph{convex}, if the white region is convex. Examples from these datasets are shown in Fig~\ref{samples}. In actual training, no preprocessing method was used. As for the specific kernels used, we used Gaussian kernels ($\kernelwa{\vctr{x}}{\vctr{y}} = e^{-\normsub{\vctr{x} - \vctr{y}}{2}/\sigma^2}$) for the kernelized models for all our experiments. To ensure that the comparisons with other models are fair, we used the regularized (two-norm regularization on weights) cross-entropy loss as the objective function for the output layer of all models. More details can be found in Appendix~\ref{settings}.

We first test the effect of using different hidden loss functions using a two-hidden-layer kMLP. The three hidden layer loss functions tested include the proposed SRS-1 loss, i.e., the SRS loss with $p=1$, the SRS-2 loss and the empirical alignment~\citep{Cristianini2002} between $\mtrxsub{G}{i}$ and $\mtrxsup{G}{\star}$, where $i$ is the hidden layer being optimized. $\mtrxsub{G}{i}$ is the kernel matrix of $k_{i+1}$ computed on $\vfnwasub{F}{S_\vctr{X}}{i}$ and $\mtrxsup{G}{\star}$ is the kernel matrix induced by $g$ on $S_\vctr{X}$. The regularization term was always chosen to be the sum of the $L^2$ norms of the weights. On \emph{convex}, this kMLP achieved a test error rate of $19.36\%$, $18.53\%$ and $21.70\%$ using alignment, SRS-2 and SRS-1 as the hidden losses, respectively. As a baseline, our best two-hidden-layer MLP achieved an error rate of $23.28\%$ on this dataset. For the rest of our experiments, we use the best result from using these three hidden losses for our greedily-trained models.


We now test the layer-wise learning algorithm against BP using the standard MNIST dataset \citep{Lecun2010}. Results from several MLPs were added as benchmarks. These models were trained with Adam or RMSProp \citep{Tieleman2012} and extra training techniques such as dropout \citep{Srivastava2014} and batch normalization (BN) \citep{Ioffe2015} were applied to boost performance. kMLPs accelerated using the proposed method ($\text{kMLP}^\text{FAST}$) were also tested, for which we randomly discarded some centers of each non-input layer before its training. Two popular acceleration methods for kernel machines were compared, including using a parametric representation ($\text{kMLP}^\text{PARAM}$), i.e., for each node in a kMLP, $\fnwa{f}{\vctr{x}} = \sum_{n=1}^{m}\alpha_n\kernelwa{\vctrsub{w}{n}}{\vctr{x}}$, $\alpha_n\cm\vctrsub{w}{n}$ learnable and $m$ a hyperparameter, and using random Fourier features ($\text{kMLP}^\text{RFF}$) \citep{Rahimi2008}.

\begin{table*}
\centering
\caption{Testing the proposed layer-wise algorithm and acceleration method on MNIST. The numbers following the model names indicate the number of hidden layers used. For $\text{kMLP}^\text{FAST}$, we also include in parentheses the ratio between the number of training examples randomly chosen as centers for the kernel machines on the layer and the size of the training set. Apart from kMLP-2 (BP), the BP kMLP results are from \citep{Zhang2017}. For this and all following tables in this paper, the entries correspond to test errors ($\%$) and $95\%$ confidence intervals ($\%$). Results with overlapping confidence intervals are considered equally good. Best results are marked in bold.}
\label{table2}
\vskip 0.1in
\centering
\begin{small}
\begin{sc}
\begin{adjustbox}{width=1\textwidth}
\begin{tabular}{lcccccr}
\toprule
\shortstack{MLP-1\\(RMSProp+BN)} & \shortstack{MLP-1\\(RMSProp+dropout)} & \shortstack{MLP-2\\(RMSProp+BN)} & \shortstack{MLP-2\\(RMSProp+dropout)} & \shortstack{kMLP-1\\(BP)} & \shortstack{kMLP-1\\(GREEDY)} & \shortstack{$\text{kMLP-1}^\text{RFF}$\\(BP)} \\
\midrule
2.05 $\pm$ 0.28 & 1.77 $\pm$ 0.26 & \textbf{1.58 $\pm$ 0.24} & 1.67 $\pm$ 0.25 & 3.44 $\pm$ 0.36 & 1.77 $\pm$ 0.26 & 2.01 $\pm$ 0.28 \\
\bottomrule
\\
\toprule
\shortstack{$\text{kMLP-1}^\text{PARAM}$\\(BP)} & \shortstack{$\text{kMLP-1}^\text{FAST}$\\(GREEDY)} & \shortstack{kMLP-2\\(BP)} & \shortstack{kMLP-2\\(GREEDY)} & \shortstack{$\text{kMLP-2}^\text{RFF}$\\(BP)} & \shortstack{$\text{kMLP-2}^\text{PARAM}$\\(BP)} & \shortstack{$\text{kMLP-2}^\text{FAST}$\\(GREEDY)} \\
\midrule
1.88 $\pm$ 0.27 & 1.75 $\pm$ 0.26 (0.54) & 3.66 $\pm$ 0.37 & \textbf{1.56 $\pm$ 0.24} & 1.92 $\pm$ 0.27 & 2.45 $\pm$ 0.30 & \textbf{1.47 $\pm$ 0.24 (1/0.19)} \\
\bottomrule
\end{tabular}
\end{adjustbox}
\end{sc}
\end{small}
\vskip 0.1in
\end{table*}

Results in Table \ref{table2} validate the effectiveness of our layer-wise algorithm. For both the single-hidden-layer and the two-hidden-layer kMLPs, the layer-wise algorithm consistently outperformed BP. The layer-wise method is also much faster than BP. In fact, it is practically impossible to use BP to train kMLP with more than two hidden layers without any acceleration method due to the computational complexity involved. Moreover, it is worth noting that the proposed acceleration trick is clearly very effective despite its simplicity and even produced models outperforming the original ones, which may be due to its regularization effect. This shows that kMLP together with the greedy learning scheme can be of practical interest even when dealing with the massive data sets in today's machine learning.

\begin{table*} 
\centering
\caption{Comparing kMLPs (trained fully layer-wise) with MLPs and other popular deep architectures trained with BP and BP enhanced by unsupervised greedy pre-training. The MLP-1 (SGD), DBN and SAE results are from \citep{Larochelle07}. Note that in order to be consistent with \citep{Larochelle07}, the MNIST results below were obtained using a train/test split (10k/50k) more challenging than what is commonly used in the literature.}
\label{table3}
\vskip 0.1in
\centering
\begin{small}
\begin{sc}
\begin{adjustbox}{width=1\textwidth}
\begin{tabular}{lcccccr}
\toprule
 & rectangles & rectangles-image & convex & mnist (50k test) & mnist (50k test) rotated & fashion-mnist \\
\midrule
MLP-1 (SGD) & 7.16 $\pm$ 0.23 & 33.20 $\pm$ 0.41 & 32.25 $\pm$ 0.41 & 4.69 $\pm$ 0.19 & 18.11 $\pm$ 0.34 & 15.47 $\pm$ 0.71 \\
MLP-1 (Adam) & 5.37 $\pm$ 0.20 & 28.82 $\pm$ 0.40 & 30.07 $\pm$ 0.40 & 4.71 $\pm$ 0.19 & 18.64 $\pm$ 0.34 & 12.98 $\pm$ 0.66 \\
MLP-1 (RMSProp+BN) & 5.37 $\pm$ 0.20 & 23.81 $\pm$ 0.37 & 28.60 $\pm$ 0.40 & 4.57 $\pm$ 0.18 & 18.75 $\pm$ 0.34 & 14.55 $\pm$ 0.69 \\
MLP-1 (RMSProp+dropout) & 5.50 $\pm$ 0.20 & 23.67 $\pm$ 0.37 & 36.28 $\pm$ 0.42 & 4.31 $\pm$ 0.18 & 14.96 $\pm$ 0.31 & 12.86 $\pm$ 0.66 \\
MLP-2 (SGD) & 5.05 $\pm$ 0.19 & \textbf{22.77 $\pm$ 0.37} & 25.93 $\pm$ 0.38 & 5.17 $\pm$ 0.19 & 18.08 $\pm$ 0.34 & 12.94 $\pm$ 0.66 \\
MLP-2 (Adam) & 4.36 $\pm$ 0.18 & 25.69 $\pm$ 0.38 & 25.68 $\pm$ 0.38 & 4.42 $\pm$ 0.18 & 17.22 $\pm$ 0.33 & \textbf{11.48 $\pm$ 0.62} \\
MLP-2 (RMSProp+BN) & 4.22 $\pm$ 0.18 & 23.12 $\pm$ 0.37 & 23.28 $\pm$ 0.37 & 3.57 $\pm$ 0.16 & 13.73 $\pm$ 0.30 & 11.51 $\pm$ 0.63 \\
MLP-2 (RMSProp+dropout) & 4.75 $\pm$ 0.19 & 23.24 $\pm$ 0.37 & 34.73 $\pm$ 0.42 & 3.95 $\pm$ 0.17 & 13.57 $\pm$ 0.30 & \textbf{11.05 $\pm$ 0.61} \\
DBN-1 & 4.71 $\pm$ 0.19 & 23.69 $\pm$ 0.37 & 19.92 $\pm$ 0.35 & 3.94 $\pm$ 0.17 & 14.69 $\pm$ 0.31 & N/A \\
DBN-3 & 2.60 $\pm$ 0.14 & \textbf{22.50 $\pm$ 0.37} & \textbf{18.63 $\pm$ 0.34} & 3.11 $\pm$ 0.15 & \textbf{10.30 $\pm$ 0.27} & N/A \\
SAE-3 & 2.41 $\pm$ 0.13 & 24.05 $\pm$ 0.37 & \textbf{18.41 $\pm$ 0.34} & 3.46 $\pm$ 0.16 & \textbf{10.30 $\pm$ 0.27} & N/A \\
kMLP-1 & \textbf{2.24 $\pm$ 0.13} & 23.29 $\pm$ 0.37 & 19.15 $\pm$ 0.34 & \textbf{3.10 $\pm$ 0.15} & 11.09 $\pm$ 0.28 & 11.72 $\pm$ 0.63 \\
$\text{kMLP-1}^\text{FAST}$ & \textbf{2.36 $\pm$ 0.13 (0.05)} & 23.86 $\pm$ 0.37 (0.01) & 20.34 $\pm$ 0.35 (0.17) & \textbf{2.95 $\pm$ 0.15 (0.1)} & 12.61 $\pm$ 0.29 (0.1) & \textbf{11.45 $\pm$ 0.62 (0.28)} \\
kMLP-2 & \textbf{2.24 $\pm$ 0.13} & 23.30 $\pm$ 0.37 & \textbf{18.53 $\pm$ 0.34} & 3.16 $\pm$ 0.15 & \textbf{10.53 $\pm$ 0.27} & \textbf{11.23 $\pm$ 0.62} \\
$\text{kMLP-2}^\text{FAST}$ & \textbf{2.21 $\pm$ 0.13 (0.3/0.3)} & 23.24 $\pm$ 0.37 (0.01/0.3) & 19.32 $\pm$ 0.35 (0.005/0.03) & 3.18 $\pm$ 0.15 (0.3/0.3) & 10.94 $\pm$ 0.27 (0.1/0.7) & \textbf{10.85 $\pm$ 0.61 (1/0.28)} \\
\bottomrule
\end{tabular}
\end{adjustbox}
\end{sc}
\end{small}
\vskip 0.1in
\end{table*}

From Table \ref{table3}, we see that the performance of kMLP is on par with some of the most popular and most mature deep architectures. In particular, the greedily-trained kMLPs compared favorably with their direct NN equivalents, i.e., the MLPs, even though neither batch normalization nor dropout was used for the former. 

\begin{figure}
\vskip 0.2in
\begin{center}
\subfloat[][Examples from test set.]{
\raisebox{.175in}{\includegraphics[width=.2\columnwidth]{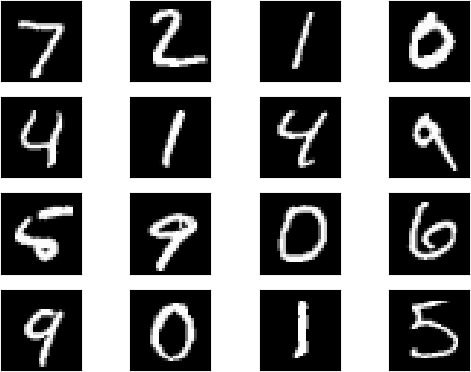}
\label{fig1a}}} 
\quad
\subfloat[][Kernel matrix of the first hidden layer (epoch 25).]{
\includegraphics[width=.2\columnwidth]{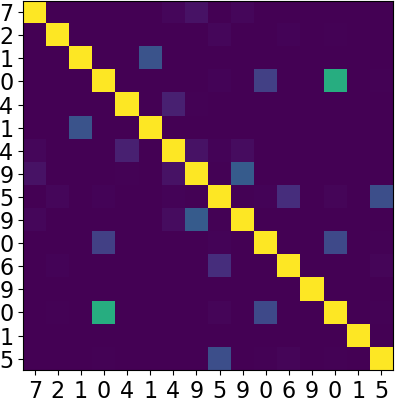}
\label{fig1b}}
\quad
\subfloat[][Kernel matrix of the second hidden layer (epoch 0).]{
\includegraphics[width=.2\columnwidth]{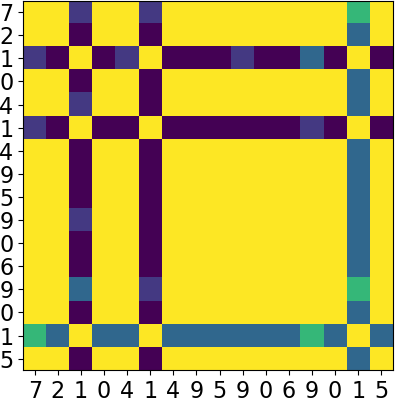}
\label{fig1c}}
\quad
\subfloat[][Kernel matrix of the second hidden layer (epoch 15).]{
\includegraphics[width=.2\columnwidth]{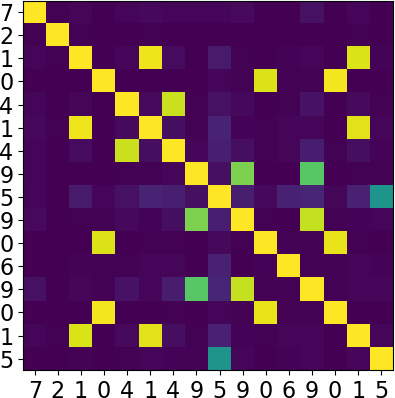}
\label{fig1d}}
\caption{Visualizing the learning dynamics in a two-hidden-layer kMLP. Each entry in the kernel matrices corresponds to the inner product between the learned representations of two examples in the RKHS. The labels are given on the two axes. The examples used to produce this figure are provided in Fig.~\ref{fig1a} in the order of the labels plotted. The darker the entry, the more distant the learned representations are in the RKHS.}
\label{hidden representation}
\end{center}
\end{figure}

In Fig.~\ref{hidden representation}, we visualize the learning dynamics within a two-hidden-layer kMLP learned layer-wise. Since by construction of the Gaussian kernel, the image vectors are all of unit norm in the RKHS, we can visualize the distance between two vectors by visualizing the value of their inner product. In Fig.~\ref{fig1d}, we can see that while the image vectors are distributed randomly prior to training (see Fig.~\ref{fig1c}), there is a clear pattern in their distribution after training that reflects the dynamics of training: the layer-wise algorithm squeezes examples from the same class closer together while pushes examples from different class farther apart. And it is easy to see that such a representation would be simple to classify. Fig.~\ref{fig1b} and \ref{fig1d} suggest that this greedy, layer-wise algorithm still learns ``deep'' representations: the higher-level representations are more distinctive for different digits than the lower-level ones. Moreover, since learning becomes increasingly simple for the upper layers as the representations become more and more well-behaved, these layers are usually easy to set up and converge very fast during training.

\subsubsection{Part 2: Kernelizing the Classic LeNet-5} 

We kernelize the output layer of the classic LeNet-5 \citep{Lecun1998} architecture and train it layer-wise with all the layers but the output layer as one layer and the output layer as a second layer. The non-output layers are trained with BP. This is to demonstrate that our kernelization method and the layer-wise algorithm are flexible in the sense that the former can be applied to only a part of the network and that the latter works well with partly-kernelized models. Since we are interested in evaluating the layer-wise algorithm on partly-kernelized NNs instead of pursuing state-of-the-art performance, we use the original LeNet-5 without increasing the size of any layer or the number of layers. ReLU \citep{Glorot2011} and max pooling were used as activations and pooling layers, respectively. Both models were optimized using Adam. The two networks were trained and tested on the unpreprocessed MNIST, Fashion-MNIST and CIFAR-10 \citep{Krizhevsky2009} datasets.

\begin{table}
\centering
\caption{Kernelizing the output layer of the classic LeNet-5. The kernelized model (kLeNet-5) was trained layer-wise.}
\label{table4}
\vskip 0.1in
\centering
\begin{small}
\begin{sc}
\begin{adjustbox}{width=.6\columnwidth}
\begin{tabular}{lccr}
\toprule
 & mnist & fashion-mnist & cifar-10 \\
\midrule
LeNet-5 & \textbf{0.76 $\pm$ 0.17} & 9.34 $\pm$ 0.57 & \textbf{36.42 $\pm$ 0.94} \\
kLeNet-5 & \textbf{0.75 $\pm$ 0.17} & \textbf{8.67 $\pm$ 0.55} & \textbf{35.87 $\pm$ 0.94} \\
\bottomrule
\end{tabular}
\end{adjustbox}
\end{sc}
\end{small}
\vskip 0.1in
\end{table}

\begin{figure}
\vskip 0.2in
\centering
\includegraphics[width=.7\linewidth]{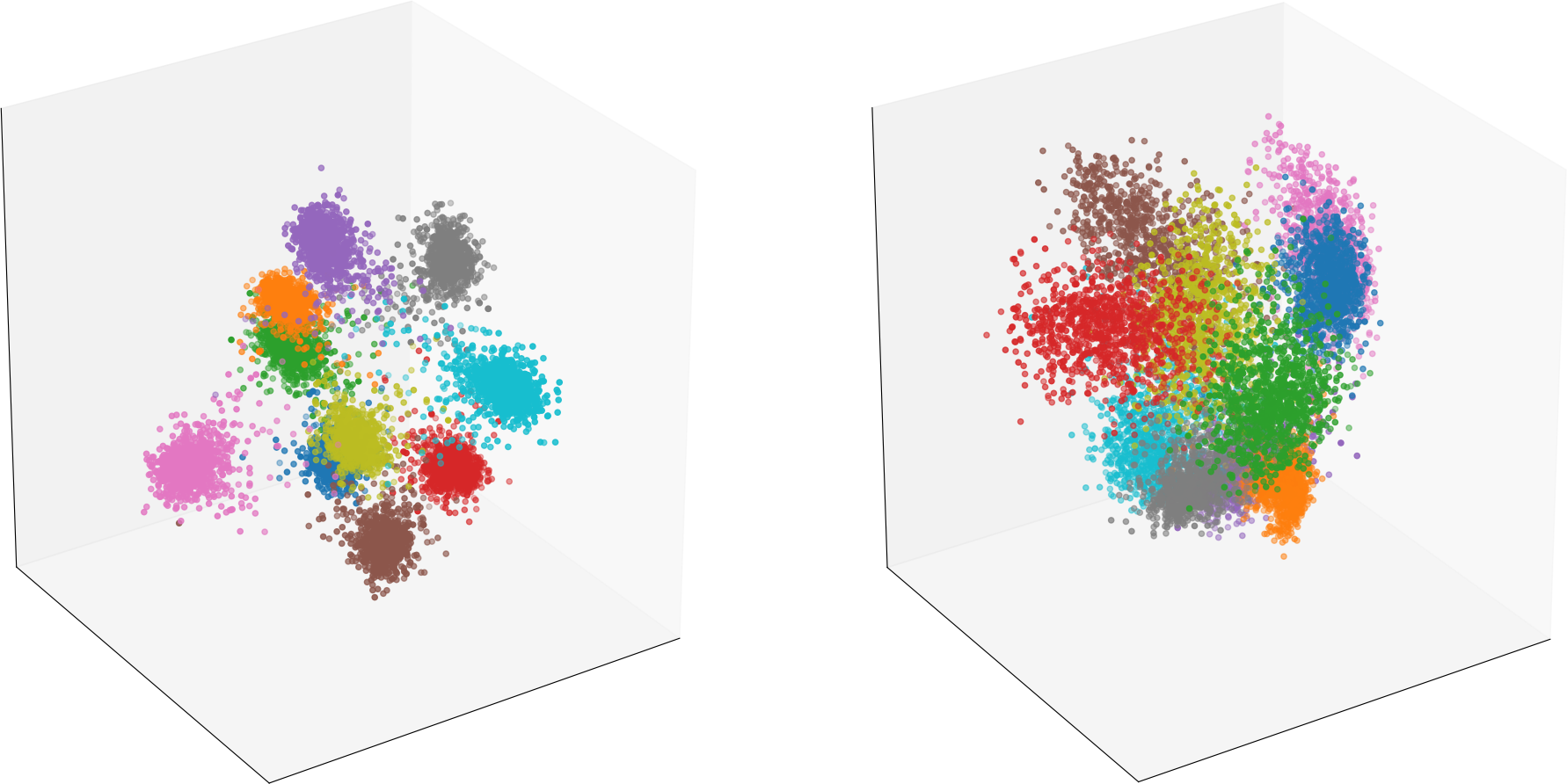}
\caption{Visualizing the data representation of the MNIST test set in the last hidden layer of kLeNet-5 (left) and LeNet-5 (right). Each color corresponds to a digit. Representations learned by kLeNet-5 are more discriminative for different digits.}
\label{hidden representation cnn}
\vskip 0.2in
\end{figure}

In Table~\ref{table4}, the results suggest that kernelization and the layer-wise algorithm resulted in marginal accuracy increase in all datasets. We emphasize that the layer-wise framework does not help the network learn intrinsically superior hypotheses compared to the traditional end-to-end methods. In that regard, it offers the same optimality guarantee as that provided by an end-to-end method such as BP. We argue that the layer-wise framework is promising because it is more light-weight and returns more information on the training of the individual layers to the user, making possible new and more flexible model selection and hyperparameter-tuning paradigms. This could serve as a tentative step toward increasing the interpretability of deep architectures. 

Fig.~\ref{hidden representation cnn} provides more insights into the difference of kLeNet-5 and LeNet-5, in which we plotted the activations of the last hidden layer of the two models after PCA dimension reduction using the MNIST test set. In particular, we see that the representations in the last hidden layer of kLetNet-5 are much more discriminative for different digits than those in the corresponding layer of LeNet-5. Note that since the two models differed only in their output layers, this observation suggests that the layer-wise training algorithm turns deep architectures into more efficient representation learners, which may prove useful for computer vision tasks that build on convolutional features \citep{gatys2015neural, gardner2015deep}. 

\section{Conclusion}
In this paper, we first proposed a family of connectionist models based on the kernel method and then presented a framework to train multilayer feedforward networks in a greedy, layer-by-layer fashion. Several realizations of the framework was provided and their optimality proven. Finally, we described a certified layer-wise training algorithm for deep feedforward architectures for classification based on the earlier realizations. Empirical results were provided to supplement out theory, in which our proposed models and the layer-wise training algorithm compared favorably with classical kernel machines as well as other popular connectionist models.

\bibliographystyle{apacite}
\bibliography{v5}
\clearpage
\begin{appendices}

\section{Experimental Setup}
\label{settings}
The data set \emph{rectangles} has 1000 training images, 200 validation images \footnote{The last 200 of the training set. Same for other datasets as well.}, and 50000 test images. The model is required to tell if a rectangle contained in an image has a larger width or length. The location of the rectangle is random. The border of the rectangle has pixel value 255 and pixels in the rest of an image all have value 0. \emph{rectangles-image} is the same as \emph{rectangles} except that the inside and outside of the rectangle are replaced by an image patch, respectively. \emph{rectangles-image} has 10000 training images, 2000 validation images, and 50000 test images. \emph{convex} consists of images in which there are white regions (pixel value 255) on black (pixel value 0) background. The model needs to tell if the region is convex. This data set has 6000 training images, 2000 validation images, and 50000 test images. \emph{mnist (50k test)} contains 10000 training images, 2000 validation images, and 50000 test images taken from the standard MNIST. \emph{mnist (50k test) rotated} is the same as the fourth except that the digits have been randomly rotated. For detailed descriptions of the data sets, see~\citep{Larochelle07}.

The experimental setup for the greedily-trained kMLPs is as follows, kMLP-1 corresponds to a one-hidden-layer kMLP with the first layer consisting of 15 to 150 kernel machines using the same Gaussian kernel and the second layer being a single or ten (depending on the number of classes) kernel machines using another Gaussian kernel. Hyperparameters were selected using the validation set. The validation set was then used in final training only for early-stopping based on validation error. For the standard MNIST and Fashion-MNIST, the last $5000$ training examples were held out as validation set. $\text{kMLP-1}^{\text{FAST}}$ is the same kMLP for which we accelerated by randomly choosing a subset of the training set as centers for the second layer after the first had been trained. The kMLP-2 and $\text{kMLP-2}^{\text{FAST}}$ are the two-hidden-layer kMLPs, the second hidden layers of which contained 15 to 150 kernel machines. Settings of all the kMLPs trained with BP can be found in \citep{Zhang2017}. Note that because it is extremely time/memory-consuming to train kMLP-2 with BP without any acceleration method, to make training possible, we could only randomly use 10000 examples from the entire training set of 55000 examples as centers for the kMLP-2 (BP) from Table~\ref{table2}. 

In Table~\ref{table3}, we compared kMLP with a one/two-hidden-layer MLP (MLP-1/MLP-2), a one/three-hidden-layer DBN (DBN-1/DBN-3) and a three-hidden-layer SAE (SAE-3). For these models, hyperparameters were also selected using the validation set. For the MLPs, the sizes of the hidden layers were chosen from the interval [25, 700]. All hyperparameters involved in Adam, RMSProp and BN were set to the suggested default values in the corresponding papers. If used, dropout or BN was added to the hidden layers and the best probability for dropout was found using the validation set. For DBN-3 and SAE-3, the sizes of the three hidden layers varied in intervals [500, 3000], [500, 4000] and [1000, 6000], respectively. DBN-1 used a much larger hidden layer than DBN-3 to obtain comparable performance. A simple calculation shows that the total numbers of parameters in the kMLPs were fewer than those in the corresponding DBNs and SAEs by orders of magnitude in all experiments. Like in the training for the kMLPs, the validation set were also reserved for early-stopping in final training. The DBNs and SAEs had been pre-trained unsupervisedly before the supervised training phase, following the algorithms described in~\citep{Hinton06b, Bengio07}. More detailed settings for these models were reported in~\citep{Larochelle07}.

\section{Proofs}
\label{proofs}
\begin{lemma}
\label{lemma33}
Suppose $\fnsub{f}{1} \in \spcsub{F}{1}\comma \ldots\comma \fnsub{f}{d} \in \spcsub{F}{d}$ are elements from sets of real-valued functions defined on $\spcsup{R}{p}$ for some $p\geq 1$, $\spc{F} \subset \spcsub{F}{1} \times \cdots \times \spcsub{F}{d}$ is a subset of their direct sum. For $\vfn{f} \in \spc{F}$, define $\omega \circ \vfn{f}: \spcsup{R}{p} \times \cdots \times \spcsup{R}{p} \times \spcsup{R}{q} \to \spc{R}$ as $\paren{\vctrsub{x}{1}\comma \ldots\comma \vctrsub{x}{m}\comma \vctr{y}} \mapsto \fnwa{\omega}{\fnwasub{f}{\vctrsub{x}{1}}{1}\comma \ldots\comma \fnwasub{f}{\vctrsub{x}{1}}{d}\comma \fnwasub{f}{\vctrsub{x}{2}}{1}\comma \ldots\comma \fnwasub{f}{\vctrsub{x}{m}}{d}\comma \vctr{y}}$, where $\vctrsub{x}{1}\comma \ldots \comma \vctrsub{x}{m} \in \spcsup{R}{p}\comma \vctr{y}\in \spcsup{R}{q}$, and $\omega : \spcsup{R}{md}\times \spcsup{R}{q} \to \spc{R}$ is bounded and $L$-Lipschitz for each $\vctr{y} \in \spcsup{R}{q}$ with respect to the Euclidean metric on $\spcsup{R}{md}$. Let $\omega \circ \spc{F} = \{\omega \circ \vfn{f}: \vfn{f}\in \spc{F}\}$. 

Define 
\begin{equation*}
\gcpxwadub{\spcsub{F}{i}}{N}{j} = \Ewasub{\sup_{f\in\spcsub{F}{i}}\frac{1}{N}\sum_{n=1}^{N}\rvarsub{Z}{n} \fnwa{f}{\rvctrdub{X}{n}{j}}}{\rvarsub{Z}{n}\comma \rvctrdub{X}{n}{j}}\comma i = 1\comma \ldots \comma d\comma j = 1\comma \ldots \comma m,
\end{equation*} 
where the $\rvctrdub{X}{n}{j}$ are i.i.d. random vectors defined on $\spcsup{R}{p}$. We have
\begin{equation} 
\label{eq123}
    \gcpxwasub{\omega \circ \spc{F}}{N} \leq 2L \sum_{i=1}^d \sum_{j=1}^m \gcpxwadub{\spcsub{F}{i}}{N}{j}.
\end{equation}
In particular, if for all $j$, the $\rvctrdub{X}{n}{j}$ upon which the Gaussian complexities of the $\spcsub{F}{i}$ are evaluated are sets of i.i.d. random vectors with the same distribution, we have $\gcpxwadub{\spcsub{F}{i}}{N}{1} = \cdots = \gcpxwadub{\spcsub{F}{i}}{N}{m} \eqqcolon \gcpxwasub{\spcsub{F}{i}}{N}$ for all $i$ and Eq. \ref{eq123} becomes
\begin{equation*}
\gcpxwasub{\omega \circ \spc{F}}{N} \leq 2mL \sum_{i=1}^d  \gcpxwasub{\spcsub{F}{i}}{N}.
\end{equation*}
\end{lemma}
This lemma is a generalization of a result on the Gaussian complexity of Lipschitz functions on $\spcsup{R}{k}$ from \citep{Bartlett2002}. And the technique used in the following proof is also adapted from there.

\begin{proof}
For the sake of brevity, we prove the case where $m = 2$. The general case uses exactly the same technique except that the notations would be more cumbersome.

Let $\spc{F}$ be indexed by $\mathcal{A}$. Without loss of generality, assume $\cardin{\mathcal{A}} < \infty$. Define 
\begin{align*}
&\rvarsub{T}{\alpha} = \sum_{n=1}^N \fnwa{\omega}{\fnwasub{f}{\rvctrsub{X}{n}}{\alpha\comma 1}\comma \ldots\comma \fnwasub{f}{\rvctrdub{X}{n}{\prime}}{\alpha\comma d}\comma \rvctrsub{Y}{n}} \rvarsub{Z}{n};\\
&\rvarsub{V}{\alpha} = L\sum_{n=1}^N \sum_{i=1}^d \paren{\fnwasub{f}{\rvctrsub{X}{n}}{\alpha\comma i}\rvarsub{Z}{n\comma i} + \fnwasub{f}{\rvctrdub{X}{n}{\prime}}{\alpha\comma i}\rvarsub{Z}{N+n\comma i}},
\end{align*}
where $\alpha \in \mathcal{A}$, $\setwcolon{\paren{\rvctrsub{X}{n}\comma \rvctrdub{X}{n}{\prime}}}{n = 1\comma\ldots\comma N}$ is a random sample of size $N$ on $\spcsup{R}{p} \times \spcsup{R}{p}$ and $\rvarsub{Z}{1}\comma \ldots\comma \rvarsub{Z}{N}\comma \rvarsub{Z}{1\comma 1}\comma \ldots\comma \rvarsub{Z}{2N\comma d}$ are i.i.d. standard normal random variables. 

Let arbitrary $\alpha\comma \beta \in \mathcal{A}$ be given, define $\normsub{\rvarsub{T}{\alpha} - \rvarsub{T}{\beta}}{2}^2 = \E \paren{\rvarsub{T}{\alpha} - \rvarsub{T}{\beta}}^2$, where the expectation is taken over the $\rvarsub{Z}{n}$. Define $\normsub{\rvarsub{V}{\alpha} - \rvarsub{V}{\beta}}{2}^2$ similarly and we have
\begin{align*}
\normsub{\rvarsub{T}{\alpha} - \rvarsub{T}{\beta}}{2}^2 &= \sum_{n=1}^N \paren{\fnwa{\omega}{\fnwasub{f}{\rvctrsub{X}{n}}{\alpha\comma 1}\comma \ldots\comma \fnwasub{f}{\rvctrdub{X}{n}{\prime}}{\alpha\comma d}\comma \rvctrsub{Y}{n}} - \fnwa{\omega}{\fnwasub{f}{\rvctrsub{X}{n}}{\beta\comma 1}\comma \ldots\comma \fnwasub{f}{\rvctrdub{X}{n}{\prime}}{\beta\comma d}\comma \rvctrsub{Y}{n}}}^2 \\
&\leq L^2\sum_{n=1}^N \sum_{i=1}^d \paren{\paren{\fnwasub{f}{\rvctrsub{X}{n}}{\alpha\comma i} - \fnwasub{f}{\rvctrsub{X}{n}}{\beta\comma i}}^2 + \paren{\fnwasub{f}{\rvctrdub{X}{n}{\prime}}{\alpha\comma i} - \fnwasub{f}{\rvctrdub{X}{n}{\prime}}{\beta\comma i}}^2}\\
&= \normsub{\rvarsub{V}{\alpha} - \rvarsub{V}{\beta}}{2}^2.
\end{align*}
By Slepian's lemma \citep{Pisier1999},
\begin{align*} 
N \egcpxwasub{\omega\circ \spc{F}}{N} &= \Esub{\rvarsub{Z}{n}} \sup_{\alpha \in \mathcal{A}} \rvarsub{T}{\alpha} \\
&\leq 2 \Esub{\rvarsub{Z}{n\comma i}\comma \rvarsub{Z}{N+n\comma i}} \sup_{\alpha \in \mathcal{A}} \rvarsub{V}{\alpha}\\
&\leq N2L \sum_{i=1}^d \paren{\egcpxwasub{\spcsub{F}{i}}{N} + \egcpxwadub{\spcsub{F}{i}}{N}{'}}.
\end{align*}
Taking the expectation of the $\rvctrsub{X}{n}\comma \rvctrdub{X}{n}{\prime}\comma \rvctrsub{Y}{n}$ on both sides proves the result.
\end{proof}

\begin{lemma}
\label{2-norm Gaussian complexity}
Given kernel $\kernel : \spcsup{R}{d_1} \times \spcsup{R}{d_1} \to \spc{R}$, let 
\begin{equation*}
\spcsub{F}{2} = \setwvert{\fn{f}: \spcsup{R}{d_1} \to \spc{R}\comma \fnwa{f}{\vctr{x}} = \sum_{\nu=1}^m\alpha_\nu \kernelwa{\vctrsub{x}{\nu}}{\vctr{x}} + b}{\boldalpha = \paren{\alpha_1\comma \ldots\comma \alpha_m} \in \spcsup{R}{m}\comma \normsub{\boldalpha}{1} \leq A\comma b\in \spc{R}},
\end{equation*} 
where the $\vctrsub{x}{\nu}$ are an $m$-subset of $S_{\rvctr{X}}$.

Define $\spcsub{F}{1} = \setwvert{\paren{\fnsub{f}{1}\comma\ldots\comma \fnsub{f}{d_1}} \colon \vctr{x} \mapsto \paren{\fnwasub{f}{\vctr{x}}{1}\comma\ldots\comma \fnwasub{f}{\vctr{x}}{d_1}}}{\vctr{x}\in\spcsup{R}{d_0}\comma \fnsub{f}{j}\in \Omega}$, where $\Omega$ is a given hypothesis class of real-valued functions on $\spcsup{R}{d_0}$. 

Also, define $$\spcsub{F}{2} \circ \spcsub{F}{1} = \setwvert{h: \vctr{x}\mapsto \sum_{\nu=1}^m \alpha_\nu \kernelwa{\vfnwa{F}{\vctrsub{x}{\nu}}}{\vfnwa{F}{\vctr{x}}} + b}{\vctr{x}\in\spcsup{R}{d_0}\comma \normsub{\boldalpha}{1} \leq A\comma b \in \spc{R}\comma \vfn{F} \in \spcsub{F}{1}}.$$ We have
\begin{equation*}
\gcpxwasub{\spcsub{F}{2} \circ \spcsub{F}{1}}{N} \leq 2ALd_1 \gcpxwasub{\Omega}{N}.
\end{equation*}
\end{lemma}
\begin{proof} 
First, note that the bias $b$ does not change $\gcpxwasub{\spcsub{F}{2} \circ \spcsub{F}{1}}{N}$. 
\begin{align*}
\egcpxwasub{\spcsub{F}{2} \circ \spcsub{F}{1}}{N} &= \E \sup_{\boldalpha\comma \vfn{F}} \frac{1}{N} \sum_{n=1}^N \sum_{\nu=1}^m \alpha_\nu \kernelwa{\vfnwa{F}{\vctrsub{x}{\nu}}}{\vfnwa{F}{\vctrsub{x}{n}}} \rvarsub{Z}{n}\\
&\leq \E \sup_{\boldalpha\comma \vfn{F} \comma \vctrsub{y}{\nu}\in\spcsup{R}{d_1}} \frac{1}{N} \sum_{n=1}^N \sum_{\nu=1}^m \alpha_\nu \kernelwa{\vctrsub{y}{\nu}}{\vfnwa{F}{\vctrsub{x}{n}}} \rvarsub{Z}{n}.
\end{align*}
Suppose the supremum over $\vctrsub{y}{\nu}$ is attained at $\rvctrsub{Y}{\nu}$, the $\rvctrsub{Y}{\nu}$ are random vectors as they are functions of the $\rvarsub{Z}{n}$. 

Write 
\begin{align*}
&\fnsub{g}{\nu}\circ \vfnwa{F}{\vctr{x}} = \kernelwa{\vfnwa{F}{\vctr{x}}}{\rvctrsub{Y}{\nu}},\\
&\omega\circ \vfnwa{F}{\vctr{x}} = \sum_{\nu=1}^m \alpha_\nu \fnsub{g}{\nu}\circ \vfnwa{F}{\vctr{x}} = \sum_{\nu=1}^m \alpha_\nu \kernelwa{\vfnwa{F}{\vctr{x}}}{\rvctrsub{Y}{\nu}}.\\
\end{align*}
Then we have 
\begin{align*}
  \egcpxwasub{\spcsub{F}{2} \circ \spcsub{F}{1}}{N} &\leq \E \sup_{\boldalpha\comma \vfn{F}} \frac{1}{N} \sum_{n=1}^N \sum_{\nu=1}^m \alpha_\nu \kernelwa{\rvctrsub{Y}{\nu}}{\vfnwa{F}{\vctrsub{x}{n}}} \rvarsub{Z}{n}\\
                                                    &= \E \sup_{\boldalpha\comma \vfn{F}} \frac{1}{N} \sum_{n=1}^N \omega\circ \vfnwa{F}{\vctr{x}} \rvarsub{Z}{n} \\&= \egcpxwasub{\omega\circ \spcsub{F}{1}}{N}.
\end{align*}
We now prove a Lipschitz property for $\omega$. For any $\boldxi_1\cm \boldxi_2\in\spcsup{R}{d_1}$, we have
\begin{align*}
\absolute{\fnwa{\omega}{\boldxi_1} - \fnwa{\omega}{\boldxi_2}} &= \absolute{\sum_{\nu=1}^m \alpha_\nu \paren{\fnwasub{g}{\boldxi_1}{\nu} - \fnwasub{g}{\boldxi_2}{\nu}}}\\
&\leq \sum_{\nu=1}^m \absolute{\alpha_\nu}\absolute{\fnwasub{g}{\boldxi_1}{\nu} - \fnwasub{g}{\boldxi_2}{\nu}}\\
&\leq A\max_\nu \absolute{\fnwasub{g}{\boldxi_1}{\nu} - \fnwasub{g}{\boldxi_2}{\nu}}\\
&= A\max_\nu \absolute{\kernelwa{\boldxi_1}{\rvctrsub{Y}{\nu}} - \kernelwa{\boldxi_2}{\rvctrsub{Y}{\nu}}}\\
&\leq A\max_\nu L_{\rvctrsub{Y}{\nu}} \normsub{\boldxi_1 - \boldxi_2}{2}\\
&\leq AL \normsub{\boldxi_1 - \boldxi_2}{2}.
\end{align*}
Therefore, $\omega\circ \vfnwa{F}{\vctr{x}}$, as a function of $\vfnwa{F}{\vctr{x}}$, is Lipschitz w.r.t. the Euclidean metric on $\spcsup{R}{d_1}$ with Lipschitz constant at most $AL$. It is easy to check that $\omega$ is bounded. Now the desired result follows from Lemma~\ref{lemma33}.
\end{proof}
\begin{proof}[\textbf{Proof of Proposition \ref{gaussian complexity kmlp}}]
The result follows from repeatedly applying Lemma \ref{2-norm Gaussian complexity}. 
\end{proof}
\begin{proof}[\textbf{Proof of Lemma \ref{equivalent def of layer-wise opt}}]
Let $\vfnsub{G}{1} = \argmin_{\vfnsub{F}{1} \in \spcdub{F}{1}{\prime}} \min_{\vfnsub{F}{2} \in \spcdub{F}{2}{\prime}} \fnwa{\tilde{R}}{\vfnsub{F}{2} \circ \vfnsub{F}{1}}$, $\vfnsub{G}{2} = \argmin_{\vfnsub{F}{2} \in \spcdub{F}{2}{\prime}} \fnwa{\tilde{R}}{\vfnsub{F}{2} \circ \vfnsub{G}{1}}$.

Suppose $\vfndub{F}{1}{\star} \neq \vfnsub{G}{1}$, 
\begin{align*}
\fnwa{\tilde{R}}{\vfnsub{G}{2} \circ \vfnsub{G}{1}} &= \min_{\vfnsub{F}{2} \in \spcdub{F}{2}{\prime}} \fnwa{\tilde{R}}{\vfnsub{F}{2}\circ \vfnsub{G}{1}} &(\text{definition of $\vfnsub{G}{2}$}) \\
&< \min_{\vfnsub{F}{2} \in \spcdub{F}{2}{\prime}} \fnwa{\tilde{R}}{\vfnsub{F}{2} \circ \vfndub{F}{1}{\star}} &(\text{definition of $\vfnsub{G}{1}$ and $\vfndub{F}{1}{\star} \neq \vfnsub{G}{1}$}) \\
&=\fnwa{\tilde{R}}{\vfndub{F}{2}{\star}\circ \vfndub{F}{1}{\star}}. &(\text{definition of $\vfndub{F}{2}{\star}$})
\end{align*}
However, this contradicts the optimality of $\vfndub{F}{2}{\star} \circ \vfndub{F}{1}{\star}$.
\end{proof}

\begin{proof}[\textbf{Proof of Theorem \ref{lemma6}}]
\label{pf lemma6}
Let $\spcdub{F}{1}{\prime}$ be the class of all $\vfndub{F}{1}{\prime}$ such that for any $\fnsub{f}{2} \in \argmin_{\fnsub{f}{2}\in \spcsub{F}{2}} \fnwa{\tilde{R}}{\fnsub{f}{2}\circ \vfndub{F}{1}{\prime}}$, there exist $\paren{\vctrsub{x}{+}\comma y_+}\comma \paren{\vctrsub{x}{-}\comma y_-} \in S$ such that $\fnwa{\ell}{\fnsub{f}{2}\circ\vfndub{F}{1}{\prime}\comma \paren{\vctrsub{x}{n}\comma y_n}} = 0$, $n = +\comma -$. 

Observe that $\vfndub{F}{1}{\star}\in \spcdub{F}{1}{\prime}$ since any $\fnsub{f}{2} \in \argmin_{\fnsub{f}{2}\in \spcsub{F}{2}} \fnwa{\tilde{R}}{\fnsub{f}{2}\circ \vfndub{F}{1}{\star}}$ is easily shown to be $\fndub{f}{2}{\star}$.

Now, suppose $\vfndub{F}{1}{\circ}$ satisfies Eq.~\ref{eq1}, if we show $\vfndub{F}{1}{\circ}\in \spcdub{F}{1}{\prime}$ and that for any $\vfndub{F}{1}{\prime}\in \spcdub{F}{1}{\prime}$, we have
\begin{equation*}
\min_{\fnsub{f}{2}\in \spcsub{F}{2}} \fnwa{\tilde{R}}{\fnsub{f}{2}\circ \vfndub{F}{1}{\prime}} \geq \min_{\fnsub{f}{2}\in \spcsub{F}{2}} \fnwa{\tilde{R}}{\fnsub{f}{2}\circ \vfndub{F}{1}{\circ}},
\end{equation*}
then by Lemma~\ref{equivalent def of layer-wise opt}, $\vfndub{F}{1}{\circ} = \vfndub{F}{1}{\star}$.

We now start the formal proof. Note that we drop the layer indices $1$ and $2$ for brevity, which will cause no confusion since the output layer will be denoted by $f$ and the input layer $\vfn{F}$. We assume that $\vfnsup{F}{\circ}$ satisfies Eq.~\ref{eq1}. Let $\fnsup{f}{\circ} \in \argmin_{\fn{f}\in \spcsub{F}{2}} \fnwa{\tilde{R}}{\fn{f}\circ \vfnsup{F}{\circ}}$. Let $\vfnsup{F}{\prime}\in \spcdub{F}{1}{\prime}$ be given and also let $\fnsup{f}{\prime} \in \argmin_{\fn{f}\in \spcsub{F}{2}} \fnwa{\tilde{R}}{\fn{f}\circ \vfnsup{F}{\prime}}$. 

\begin{claim}
\label{claim1}
$$\normsub{\fnwa{\phi}{\vctr{x}}}{H} = \sqrt{c}\comma \forall \vctr{x}\in \spcsup{R}{d_1}.$$
\end{claim}
\begin{subproof}[\textbf{Proof of Claim~\ref{claim1}}]
$$c = \kernelwa{\vctr{x}}{\vctr{x}} = \innersub{\fnwa{\phi}{\vctr{x}}}{\fnwa{\phi}{\vctr{x}}}{H} = \normsub{\fnwa{\phi}{\vctr{x}}}{H}^2,$$
which implies $\normsub{\fnwa{\phi}{\vctr{x}}}{H} = \sqrt{c}$.
\end{subproof}

\begin{claim}
\label{claim2}
\begin{align*}
&\fnwa{\phi}{\vfnwasup{F}{\vctr{x}}{\circ}} = \fnwa{\phi}{\vfnwasup{F}{\vctrsup{x}{\prime}}{\circ}}\comma \quad \forall \vctr{x}\comma \vctrsup{x}{\prime}\in S_{\rvctr{X}} \text{ with } y = y';\\
&\fnwa{\phi}{\vfnwasup{F}{\vctrsub{x}{+}}{\circ}} = \fnwa{\phi}{\vfnwasup{F}{\vctrsub{x}{-}}{\circ}}\comma \quad \forall \vctrsub{x}{+}\comma \vctrsub{x}{-}\in S_{\rvctr{X}}.\\
\end{align*}
\end{claim}
\begin{subproof}[\textbf{Proof of Claim~\ref{claim2}}]
By Cauchy-Schwarz inequality and Claim~\ref{claim1}, 
\begin{equation*}
0 < c = \kernelwa{\vfnwasup{F}{\vctr{x}}{\circ}}{\vfnwasup{F}{\vctrsup{x}{\prime}}{\circ}} = \innersub{\fnwa{\phi}{\vfnwasup{F}{\vctr{x}}{\circ}}}{\fnwa{\phi}{\vfnwasup{F}{\vctrsup{x}{\prime}}{\circ}}}{H} \leq \normsub{\fnwa{\phi}{\vfnwasup{F}{\vctr{x}}{\circ}}}{H}\normsub{\fnwa{\phi}{\vfnwasup{F}{\vctrsup{x}{\prime}}{\circ}}}{H} = c.
\end{equation*}
So the equality holds in Cauchy-Schwarz and we have $\fnwa{\phi}{\vfnwasup{F}{\vctr{x}}{\circ}} = p\fnwa{\phi}{\vfnwasup{F}{\vctrsup{x}{\prime}}{\circ}}$ for some $p>0$. Again by Claim~\ref{claim1}, $p=1$.

The second part of this claim follows from $\kernelwa{\vfnwasup{F}{\vctrsub{x}{+}}{\circ}}{\vfnwasup{F}{\vctrsub{x}{-}}{\circ}} = a \neq c$.
\end{subproof}

\begin{claim}
\label{claim3}
For any $\vctrsub{x}{+}\cm \vctrsub{x}{-}\in S_{\rvctr{X}}$ and any $\vfn{F}: \spcsup{R}{d_0} \to \spcsup{R}{d_1}$,
\begin{align*}
    \sqrt{2(c - a)} = \normsub{
        \fnwa{\phi}{\vfnwasup{F}{\vctrsub{x}{+}}{\circ}}
        -\fnwa{\phi}{\vfnwasup{F}{\vctrsub{x}{-}}{\circ}} 
        }{
        H
        } \geq \normsub{
        \fnwa{\phi}{\vfnwa{F}{\vctrsub{x}{+}}}
        -\fnwa{\phi}{\vfnwa{F}{\vctrsub{x}{-}}} 
        }{
        H
    }.
\end{align*}
\end{claim}
\begin{subproof}[\textbf{Proof of Claim~\ref{claim3}}]
    \begin{align*}
        &\normsub{
        \fnwa{\phi}{\vfnwasup{F}{\vctrsub{x}{+}}{\circ}}
        -\fnwa{\phi}{\vfnwasup{F}{\vctrsub{x}{-}}{\circ}} 
        }{
        H
    }^2 \\ 
        &\quad= 
        \normsub{
        \fnwa{\phi}{\vfnwasup{F}{\vctrsub{x}{+}}{\circ}}
        }{
        H
        }^2 + 
        \normsub{
        \fnwa{\phi}{\vfnwasup{F}{\vctrsub{x}{-}}{\circ}}
        }{
        H
        }^2 - 
        2\innersub{
            \fnwa{\phi}{\vfnwasup{F}{\vctrsub{x}{+}}{\circ}} 
            }{
            \fnwa{\phi}{\vfnwasup{F}{\vctrsub{x}{-}}{\circ}} 
            }{
            H
        } \\
      &\quad= 
      c + c - 2\kernelwa{
          \vfnwasup{F}{\vctrsub{x}{+}}{\circ} 
          }{
          \vfnwasup{F}{\vctrsub{x}{-}}{\circ} 
      } \\
      &\quad= 2c - 2a \\
      &\quad\geq 2c - 2\kernelwa{
          \vfnwa{F}{\vctrsub{x}{+}} 
          }{
          \vfnwa{F}{\vctrsub{x}{-}} 
      } \\
      &\quad= 
        \normsub{
        \fnwa{\phi}{\vfnwa{F}{\vctrsub{x}{+}}}
        }{
        H
        }^2 + 
        \normsub{
        \fnwa{\phi}{\vfnwa{F}{\vctrsub{x}{-}}}
        }{
        H
        }^2 - 
        2\innersub{
            \fnwa{\phi}{\vfnwa{F}{\vctrsub{x}{+}}} 
            }{
            \fnwa{\phi}{\vfnwa{F}{\vctrsub{x}{-}}} 
            }{
            H
} \\
      &\quad= \normsub{
        \fnwa{\phi}{\vfnwa{F}{\vctrsub{x}{+}}}
        -\fnwa{\phi}{\vfnwa{F}{\vctrsub{x}{-}}} 
        }{
        H
    }^2
\end{align*}
\end{subproof}

\begin{claim}
    \label{claim4}
    $$\fnwa{\tilde{R}}{f^\circ \circ \vfnsup{F}{\circ}} = \tau \normsub{
        \vctrsub{w}{f^\circ}
        }{
        H
        } = \frac{
        2\tau
        }{
        \normsub{
        \fnwa{\phi}{\vfnwasup{F}{\vctrsub{x}{+}}{\circ}}
        -\fnwa{\phi}{\vfnwasup{F}{\vctrsub{x}{-}}{\circ}} 
        }{
        H
    }
  }\cm\forall\vctrsub{x}{+}\cm\vctrsub{x}{-}\in S_\vctr{X}.$$
\end{claim}
\begin{subproof}[\textbf{Proof of Claim~\ref{claim4}}]
    \begin{align*}
        &\fnwa{\tilde{R}}{f^\circ \circ \vfnsup{F}{\circ}}\\
        &\quad=
        \frac{1}{N} \sum_{n=1}^{N} \max \paren{
            0\cm 1 - y_n \fnwasup{
                f
                }{
                \vfnwasup{
                    F
                    }{
                    \vctrsub{x}{n}
                    }{
                    \circ
                }
                }{
                \circ
            }
            } + \tau \normsub{
            \vctrsub{w}{f^\circ}
            }{
            H
        } \\
        &\quad= 
        \kappa\max\paren{
        0\cm 1 - y_{+}
        \fnwasup{
            f
            }{
            \vfnwasup{
                F
                }{
                \vctrsub{x}{+}
                }{
                \circ
            }
            }{
            \circ
        }
    } + 
    (1 -
    \kappa)\max\paren{
        0\cm 1 - y_{-}
        \fnwasup{
            f
            }{
            \vfnwasup{
                F
                }{
                \vctrsub{x}{-}
                }{
                \circ
            }
            }{
            \circ
        }
    } + \tau \normsub{
                \vctrsub{w}{f^\circ}
            }{
            H
        },\\
    \end{align*}
   for any pair of $\vctrsub{x}{+}\cm \vctrsub{x}{-} \in S_{\rvctr{X}}$. Let $$\zeta_{f^\circ} = y_+ \fnwasup{
        f
        }{
        \vfnwasup{
            F
            }{
            \vctrsub{x}{+}
            }{
            \circ
            }    
        }{
            \circ
            } + y_- \fnwasup{
            f
            }{
            \vfnwasup{
                F
                }{
                \vctrsub{x}{-}
                }{
                \circ
            }
            }{
            \circ
            } = \normsub{
            \vctrsub{w}{f^\circ}
            }{
            H
        }
        \normsub{
        \fnwa{\phi}{\vfnwasup{F}{\vctrsub{x}{+}}{\circ}}
        -\fnwa{\phi}{\vfnwasup{F}{\vctrsub{x}{-}}{\circ}} 
        }{
        H
    } \cos \theta_{f^\circ}.$$ We have
    \begin{align*}
        \fnwa{\tilde{R}}{f^\circ \circ \vfnsup{F}{\circ}} = 
        \kappa \max (0\cm 1 - t_{f^\circ}) + (1 - \kappa) \max (0\cm 1 -
        (\zeta_{f^\circ}
        - t_{f^\circ})) + \tau\normsub{
            \vctrsub{w}{f^\circ}
            }{
            H
        },
    \end{align*}
    where $t_{f^\circ} = \fnwasup{
        f
        }{
        \vfnwasup{
            F
            }{
            \vctrsub{x}{+}
            }{
            \circ
        }
        }{
        \circ
    }$.

    Note that by definition of $f^\circ$, 
    \begin{align*}
        \fnwa{\tilde{R}}{f^\circ \circ
            \vfnsup{F}{\circ}} &= \min_{f}\fnwa{\tilde{R}}{f \circ
        \vfnsup{F}{\circ}}\\ 
                          &= \min_{\zeta_f\cm t_f\cm
    \normsub{\vctrsub{w}{f}}{H}} \kappa \max (0\cm 1 - t_{f}) + (1 - \kappa) \max (0\cm 1 -
        (\zeta_{f}
        - t_{f})) + \tau\normsub{
            \vctrsub{w}{f}
            }{
            H
        }.
    \end{align*}

There are four possible cases that the terms inside the minimum operator can be simplified to:
\begin{enumerate}[label=(\arabic*)]
\item\label{claim4_1} If $1\geq t_f \geq \zeta_{f} - 1\cm \zeta_f\leq
    2$, to $1 - (1-\kappa)
        \zeta_f + (1 - 2\kappa) t_f + \tau\normsub{
            \vctrsub{w}{f}
            }{
            H
        }$;
    \item\label{claim4_2} If $t_f\geq \max(1\cm \zeta_f - 1)$, to $(1-\kappa)(1 - \zeta_f + t_f) + \tau\normsub{
            \vctrsub{w}{f}
            }{
            H
        }$;
    \item\label{claim4_3} If $t_f\leq \min(1\cm \zeta_f - 1)$, to $\kappa(1 - t_f)+ \tau\normsub{
            \vctrsub{w}{f}
            }{
            H
        }$;
    \item\label{claim4_4} If $1\leq t_f\leq \zeta_f - 1\cm \zeta_f\geq 2$, to $\tau\normsub{
            \vctrsub{w}{f}
            }{
            H
        }$.
\end{enumerate}

If $\zeta_f \geq 2$, for each fixed $\zeta_f\cm \normsub{\vctrsub{w}{f}}{H}\cm
t_f$, we have that the values of $\tilde{R}$ in \ref{claim4_2}, \ref{claim4_3} are no less
than that in \ref{claim4_4} and that their minima agree. Therefore, when $\zeta_f\geq 2$, $\fnwa{\tilde{R}}{f^\circ
    \circ \vfnsup{F}{\circ}} = \tau\normsub{
            \vctrsub{w}{f^\circ}
            }{
            H
        }$.

        On the other hand, if $\zeta_f\leq 2$, then for each fixed $\zeta_f\cm
        \normsub{
            \vctrsub{w}{f}
            }{
            H
        }$, first note $t_f\in \spc{R}$ can be chosen freely by adjusting $b$.
        Also, since $\max(1\cm \zeta_f - 1) = 1$ and $\min(1\cm \zeta_f - 1) =
        \zeta - 1$, by working out the minima over $t_f$ in \ref{claim4_1},
        \ref{claim4_2}, and \ref{claim4_3}, respectively, we have $\fnwa{\tilde{R}}{f^\circ
            \circ \vfnsup{F}{\circ}} = \min(\kappa\cm 1 - \kappa)(2 -
            \zeta_{f^\circ}) + \tau \normsub{
                \vctrsub{w}{f^\circ}
                }{
                H
            }$.

            Note that we have $\zeta_{f^\circ} = \normsub{
                \vctrsub{w}{f^\circ}
                }{
                H
            } \psi^\circ \cos \theta_{f^\circ}$, where $\psi^\circ = \normsub{
        \fnwa{\phi}{\vfnwasup{F}{\vctrsub{x}{+}}{\circ}}
        -\fnwa{\phi}{\vfnwasup{F}{\vctrsub{x}{-}}{\circ}} 
        }{
        H
    }$, we can rewrite the earlier result in terms of $\zeta_{f^\circ}$ and
    $\cos \theta_{f^\circ}$. Consequently, we now determine the minimum over
    $\zeta_{f}$ and $\cos \theta_f$ of the resulting expression.

    To this end, first observe that for each $\zeta_f$, one can choose $\cos
    \theta_f \in \lcrc{-1}{1}$ freely by adjusting
    $\normsub{\vctrsub{w}{f}}{H}$ under the constraint that the two quantities
    must be of the same sign, if both are nonzero. Therefore, for each
    $\zeta_{f}\geq
    2$, $$\min_{
    \cos \theta_f} \fnwa{\tilde{R}}{f \circ \vfnsup{F}{\circ}} =
    \fnwa{\tilde{R}}{f \circ \vfnsup{F}{\circ}}\mid_{\cos
    \theta_f = 1} = \frac{\tau \zeta_{f}}{\psi^\circ}.$$
    Similarly, for each $\zeta_{f}\leq 2$, we have $\min_{
    \cos \theta_f} \fnwa{\tilde{R}}{f \circ \vfnsup{F}{\circ}} = \min(\kappa\cm 1 - \kappa)(2 - \zeta_{f})
    + \tau \absolute{\zeta_{f}}/\psi^\circ$.

    Combining these two cases and using the assumption on $\tau$, it is easy to see that $\fnwa{\tilde{R}}{f^\circ
    \circ \vfnsup{F}{\circ}} = \min_{\zeta_{f}}\fnwa{\tilde{R}}{f \circ
\vfnsup{F}{\circ}} = \fnwa{\tilde{R}}{f \circ
\vfnsup{F}{\circ}}\mid_{\zeta_f = 2} = 2\tau/\psi^\circ$. This proves the claim.
\end{subproof}

\rmk{
    By Claim~\ref{claim4}, $\vfnsup{F}{\circ}\in \spcdub{F}{1}{'}$.
}

\begin{claim}
    \label{claim5}
    For any $\vfnsup{F}{'}\in \spcdub{F}{1}{'}\cm \min_{f\in \spcsub{F}{2}} 
    \fnwa{
        \tilde{R}
    }{
        f\circ \vfnsup{F}{'}
    } \geq 
    \min_{f\in \spcsub{F}{2}} 
    \fnwa{
        \tilde{R}
    }{
        f\circ \vfnsup{F}{\circ}
    }$.
\end{claim}
\begin{subproof}[\textbf{Proof of Claim~\ref{claim5}}]
By Claim~\ref{claim4}, it amounts to prove $$\fnwa{\tilde{R}}{f'\circ \vfnsup{F}{'}}
\geq \frac{2\tau}{\normsub{
        \fnwa{\phi}{\vfnwasup{F}{\vctrsub{x}{+}}{\circ}}
        -\fnwa{\phi}{\vfnwasup{F}{\vctrsub{x}{-}}{\circ}} 
        }{
        H
    }},$$
for an arbitrary pair of $\vctrsub{x}{+}\cm \vctrsub{x}{-}\in S_\rvctr{X}$. Suppose $(\vctrdub{x}{+}{'}\cm y'_+)\cm (\vctrdub{x}{-}{'}\cm y'_-)$ are a pair of examples
with $\vctrdub{x}{+}{'}\cm \vctrdub{x}{-}{'}\in S_\rvctr{X}$, and $\fnwa{\ell}{f'\circ \vfnsup{F}{'}\cm (\vctrdub{x}{n}{'}\cm y'_n)} = 0\cm n = +\cm -$, then we have $y'_+ \fnwasup{f}{\vctrdub{x}{+}{'}}{'} + y'_- \fnwasup{f}{\vctrdub{x}{-}{'}}{'}\geq 2$.

Since $y'_+ \fnwasup{f}{\vctrdub{x}{+}{'}}{'} + y'_- \fnwasup{f}{\vctrdub{x}{-}{'}}{'} = \normsub{\vctrsub{w}{f'}}{H} \normsub{
        \fnwa{\phi}{\vfnwasup{F}{\vctrdub{x}{+}{'}}{'}}
        -\fnwa{\phi}{\vfnwasup{F}{\vctrdub{x}{-}{'}}{'}} 
        }{
        H
    } \cos\theta_{f'}$, it is implied that $\cos\theta_{f'} \in \lorc{0}{1}\cm \normsub{
        \fnwa{\phi}{\vfnwasup{F}{\vctrdub{x}{+}{'}}{'}}
        -\fnwa{\phi}{\vfnwasup{F}{\vctrdub{x}{-}{'}}{'}} 
        }{
        H
    } > 0$ and $\normsub{\vctrsub{w}{f'}}{H} \geq 2/\normsub{
        \fnwa{\phi}{\vfnwasup{F}{\vctrdub{x}{+}{'}}{'}}
        -\fnwa{\phi}{\vfnwasup{F}{\vctrdub{x}{-}{'}}{'}} 
        }{
        H
    }$. Therefore, 
\begin{align*}
  \fnwa{\tilde{R}}{f'\circ \vfnsup{F}{'}} &\geq \tau\normsub{\vctrsub{w}{f'}}{H}\\
                                          &\geq \frac{2\tau}{\normsub{
        \fnwa{\phi}{\vfnwasup{F}{\vctrdub{x}{+}{'}}{'}}
        -\fnwa{\phi}{\vfnwasup{F}{\vctrdub{x}{-}{'}}{'}} 
        }{
        H
  }}\\ &\geq \frac{2\tau}{\normsub{
        \fnwa{\phi}{\vfnwasup{F}{\vctrdub{x}{+}{'}}{\circ}}
        -\fnwa{\phi}{\vfnwasup{F}{\vctrdub{x}{-}{'}}{\circ}} 
        }{
        H
    }}\\
       &=\fnwa{\tilde{R}}{f^\circ\circ\vfnsup{F}{\circ}}.
\end{align*}
\end{subproof}

This concludes the proof of the theorem.
\end{proof}

\begin{proof}[\textbf{Proof of Theorem~\ref{new theorem 4.5}}]
Denote with $\spcdub{F}{2}{\prime}$ the set of all $\vfndub{F}{2}{\prime}$ such that for all $j$, $
    \normsub{
        \vctrsub{w}{f_2^{j\prime}}
        }{
        H
    }>0$.

    Denote with $\spcdub{F}{1}{\prime}$ the set of all $\vfndub{F}{1}{\prime}$ such that for any $\vfnsub{F}{2}\in 
        \argmin_{
            \vfnsub{F}{2}\in\spcdub{F}{2}{\prime}
        }
        \fnwa{
            \tilde{R}
            }{
    \vfnsub{F}{2}\circ\vfndub{F}{1}{\prime}
            }$, $\vfnsub{F}{2}$ satisfies:$$
    \exists 
    (\vctrsub{x}{+}\cm y_+)\cm
    (\vctrsub{x}{-}\cm y_-)
    \in S_{\rvctr{X}}\times S_\rvar{Y}\st
    \fnwa{
        \ell
        }{
\vfnsub{F}{2}\circ\vfndub{F}{1}{\prime}\cm
        (\vctrsub{x}{+}\cm y_+)\cm
        (\vctrsub{x}{-}\cm y_-)
        }= 0.
    $$

    Using the same argument as in the beginning of the proof of Theorem~\ref{lemma6}, we have $
    \vfndub{
        F
	}{
	1
	}{
        \star
    }\in
    \spcdub{
        F
        }{
        1
        }{
        '
    }$. Let $
    \vfndub{
        F
	}{
	1
	}{
        \prime
    }\in
    \spcdub{
        F
        }{
        1
        }{
       \prime 
    }
    $
    be given and suppose $
    \vfndub{
        F
	}{
	1
	}{
        \circ
    }
    $
    satisfies Eq.~\ref{eq100}. Let $$
    \vfndub{
        F
	}{
	2
	}{
        \prime
    }\in
        \argmin_{
            \vfnsub{F}{2}\in\spcdub{F}{2}{\prime}
        }
        \fnwa{
            \tilde{R}
            }{
    \vfnsub{F}{2}\circ \vfndub{F}{1}{\prime}
        }, \text{ and } 
    \vfndub{
        F
	}{
	2
	}{
        \circ
    }\in
        \argmin_{
            \vfnsub{F}{2}\in\spcdub{F}{2}{\prime}
        }
        \fnwa{
            \tilde{R}
            }{
    \vfnsub{F}{2}\circ \vfndub{F}{1}{\circ}
        }.
    $$
    Then by Lemma~\ref{equivalent def of layer-wise opt}, the proof is complete if we can show $
    \fnwa{
        \tilde{R}
        }{
        \vfndub{
            F
	    }{
	    2
	    }{
            \prime
        }\circ
        \vfndub{
            F
	    }{
	    1
	    }{
            \prime
        }
        }\geq
    \fnwa{
        \tilde{R}
        }{
        \vfndub{
            F
	    }{
	    2
	    }{
            \circ
        }\circ
        \vfndub{
            F
	    }{
	    1
	    }{
            \circ
        }
        }
    $.

    To this end, first note that Claims~\ref{claim1}, \ref{claim2}, \ref{claim3} from the proof of Theorem~\ref{lemma6} evidently hold here as well. Define $\psi = 1/N^2 \sum_{n\cm m=1}^N \ind{y_m\neq y_n}$.

    \begin{claim}
        \label{claim6}
        $
        \vfnwadub{
            F
            }{
            \vctr{x}
            }{
    2}{\circ
        } =
        \vfnwadub{
            F
            }{
            \vctrsup{x}{\prime}
            }{
    2}{\circ
        }, \forall \vctr{x}\cm\vctrsup{x}{\prime}\in S_{\rvctr{X}} 
        \text{ with } y = y^\prime
        $.
    \end{claim}

    \begin{subproof}[\bf{Proof of Claim~\ref{claim6}}]
       $
        \forall \vctr{x}\cm\vctrsup{x}{\prime}\in S_{\rvctr{X}}\cm
        $
        \begin{align*}
             \vfnwadub{
                 F
                 }{
                 \vctr{x}
                 }{
	 2}{\circ
             }-
             \vfnwadub{
                 F
                 }{
                 \vctrsup{x}{\prime}
                 }{
	 2}{\circ
             }&= 
             \paren{
                 \fnwadub{
                     f
                     }{
                     \vctr{x}
                     }{
                     2
                     }{
                     1\circ
                 }-
                 \fnwadub{
                     f
                     }{
                     \vctrsup{x}{\prime}
                     }{
                     2
                     }{
                     1\circ
                 }\cm \ldots
            }\\&=
             \paren{
                 \innersub{
                     \vctrsub{w}{f_2^{1\circ}}
                     }{
                     \fnwa{
                         \phi
                         }{
		 \vfnwadub{F}{\vctr{x}}{1}{\circ}
                         }-
                     \fnwa{
                         \phi
                         }{
		 \vfnwadub{F}{\vctrsup{x}{\prime}}{1}{\circ}
                         }
                     }{
                     H
                 }\cm \ldots
             }=
             (0\cm\ldots),
        \end{align*}
        where we have used Claim~\ref{claim2} for the last equality.
    \end{subproof}

    Combining this claim with our earlier assumptions on $h$, we can simplify the objective function
    $$
    \fnwa{
        \tilde{R}
        }{
        \vfndub{
            F
	    }{
	    2
	    }{
            \circ
        }\circ
        \vfndub{
            F
	    }{
	    1
	    }{
            \circ
        }
        }=\psi
        \paren{
            \fnwa{
                h
                }{
                \normsub{
                    \vfnwadub{
                        F
                        }{
                        \vctrsub{x}{+}
                        }{
		2}{\circ
                    }-
                    \vfnwadub{
                        F
                        }{
                        \vctrsub{x}{-}
                        }{
		2}{\circ
                    }
                }{
                q
                }
                }
            -b
        }^p + 
        \tau
        \fnwa{
        t
        }{
        \normsub{\vctrsub{w}{f_2^{1\circ}}}{H}\cm\ldots\cm
        \normsub{\vctrsub{w}{f_2^{d_2\circ}}}{H}
    },
    $$
    where $\vctrsub{x}{+}\cm\vctrsub{x}{-}\in S_\rvctr{X}$ are arbitrary.

    Rewrite the above expression as
    \begin{align*}
    &\fnwa{
        \tilde{R}
        }{
        \vfndub{
            F
	    }{
	   2 
	    }{
            \circ
        }\circ
        \vfndub{
            F
	    }{
	    1
	    }{
            \circ
        }
        }\\&\quad=\psi
        \paren{
            \fnwa{
                h
                }{
                \paren{
                    \sum_j^{d_2}
                    \normsub{
                        \vctrsub{w}{f_2^{j\circ}}
                        }{
                        H
                    }^q
                    \normsub{
                        \fnwa{
                            \phi
                        }{
                           \vfnwadub{
                               F
                               }{
                               \vctrsub{x}{+}
                               }{
		       1}{\circ
                           }
                           }
                       - 
                       \fnwa{
                           \phi
                       }{
                           \vfnwadub{
                               F
                               }{
                               \vctrsub{x}{-}
                               }{
		       1}{\circ
                           }
                           }
               }{
                        H
                    }^q
                    \paren{
                        \cos \theta_{f_2^{j\circ}}
                    }^q
                }^{1/q}
                }
            -b
    }^p\\ &\quad+ 
        \fnwa{
        t
        }{
        \normsub{\vctrsub{w}{f_2^{1\circ}}}{H}\cm\ldots\cm
        \normsub{\vctrsub{w}{f_2^{d_2\circ}}}{H}
    } 
    \end{align*}

    \begin{claim}
        \label{claim7}
        $\paren{\cos \theta_{f_2^{j\circ}}}^2 = 1\cm \forall j$.
    \end{claim}

    \begin{subproof}[\bf{Proof of Claim~\ref{claim7}}]
        This claim follows from noting that for each $
        \normsub{
            \vctrsub{
                w
                }{
                f_2^j
            }
            }{
            H
        }
        $, $
        \paren{
            \cos \theta_{f_2^j}
        }^2
        $ may be chosen freely and since the $
        \normsub{
            \vctrsub{
                w
                }{
                f_2^j
            }
            }{
            H
        }
        $ are nonzero by the definition of $\spcdub{F}{2}{\prime}$, it is easy to see that the unique minimizers of the $
        \paren{
            \cos \theta_{f_2^j}
        }^2
        $ are $
        \paren{
            \cos \theta_{f_2^j}
        }^2 = 1\cm \forall j
        $.
    \end{subproof}

    Using Claim~\ref{claim3} and the above claim, we further simplify the objective function into 
    \begin{align*}
    &\fnwa{
        \tilde{R}
        }{
        \vfndub{
            F
	    }{
	    2
	    }{
            \circ
        }\circ
        \vfndub{
            F
	    }{
	    1
	    }{
            \circ
        }
        }\\&\quad=\psi
        \paren{
            \fnwa{
                h
                }{
                \sqrt{
                    2(c-a)
                }
                \paren{
                    \sum_j^{d_2}
                    \normsub{
                        \vctrsub{w}{f_2^{j\circ}}
                        }{
                        H
                    }^q
                    }^{1/q}
		    }
                -b
    }^p+ 
        \tau
        \fnwa{
        t
        }{
        \normsub{\vctrsub{w}{f_2^{1\circ}}}{H}\cm\ldots\cm
        \normsub{\vctrsub{w}{f_2^{d_2\circ}}}{H}
    }\\
     &\quad=\min_{\vctrsub{w}{f_2^j}}
    \psi
        \paren{
            \fnwa{
                h
                }{
                \sqrt{
                    2(c-a)
                }
                \paren{
                    \sum_j^{d_2}
                    \normsub{
                        \vctrsub{w}{f_2^j}
                        }{
                        H
                    }^q
                    }^{1/q}
                    }
                -b
    }^p+ 
        \tau
        \fnwa{
        t
        }{
        \normsub{\vctrsub{w}{f_2^1}}{H}\cm\ldots\cm
        \normsub{\vctrsub{w}{f_2^{d_2}}}{H}
    } 
    \end{align*}

    Now, let $
    \paren{
        \vctrdub{
            x
            }{
            +
            }{
            \prime
        }\cm
        \vctrdub{
            x
            }{
            - 
            }{
            \prime
        }
    }\in
    \setonly{
        \argmax_{
            \vctrsub{x}{+}\cm
            \vctrsub{x}{-}\in S_{\rvctr{X}}
        }
        \normsub{
            \vfnwadub{
                F
                }{
                \vctrsub{x}{+}
		}{
		2
		}{
                \prime
            }-
            \vfnwadub{
                F
                }{
                \vctrsub{x}{-}
		}{
		2
		}{
                \prime
            }
            }{
            q
        }
    }
    $, we have
    \begin{align*}
        &\fnwa{
            \tilde{R}
            }{
    \vfndub{F}{2}{\prime}\circ\vfndub{F}{1}{\prime}
            }\\
        &\quad\geq
        \psi
        \paren{
        \fnwa{
            h
            }{
            \normsub{
                \vfnwadub{
                    F
                }{
                \vctrdub{x}{+}{\prime}
                }{
	2}{\prime
            } -
                \vfnwadub{
                    F
                }{
                \vctrdub{x}{-}{\prime}
                }{
	2}{\prime
            }
                }{
                q
            }
            }
     - b}^p
    +
    \tau
    \fnwa{
        t
        }{
        \normsub{\vctrsub{w}{f_2^{1\prime}}}{H}\cm\ldots\cm
        \normsub{\vctrsub{w}{f_2^{d_2\prime}}}{H}
    }\\
         &\quad\geq
         \psi
        \paren{
        \fnwa{
            h
            }{
            \normsub{
                \fnwa{
                    \phi
                    }{
                    \vfnwadub{
                        F
                        }{
                        \vctrdub{x}{+}{\prime}
                        }{
		1}{\prime
                    }
                    }
                -
                \fnwa{
                    \phi
                    }{
                    \vfnwadub{
                        F
                        }{
                        \vctrdub{x}{-}{\prime}
                        }{
		1}{\prime
                    }
                    }
                }{
                H
            }
            \paren{
                \sum_{j=1}^{d_2}
                \normsub{
                    \vctrsub{w}{f_2^{j\prime}}
                    }{
                    H
                }^q
            }^{1/q}
            }
     - b}^p\\
         &\quad\quad+
    \tau
    \fnwa{
        t
        }{
        \normsub{\vctrsub{w}{f_2^{1\prime}}}{H}\cm\ldots\cm
        \normsub{\vctrsub{w}{f_2^{d_2\prime}}}{H}
    }\\
    &\quad\geq \psi
        \paren{
        \fnwa{
            h
            }{
            \sqrt{2(c-a)}
            \paren{
                \sum_{j=1}^{d_2}
                \normsub{
                    \vctrsub{w}{f_2^{j\prime}}
                    }{
                    H
                }^q
            }^{1/q}
            }
     - b}^p+
    \tau
    \fnwa{
        t
        }{
        \normsub{\vctrsub{w}{f_2^{1\prime}}}{H}\cm\ldots\cm
        \normsub{\vctrsub{w}{f_2^{d_2\prime}}}{H}
    }\\
    &\quad\geq 
    \fnwa{
        \tilde{R}
        }{
\vfndub{F}{2}{\circ}\circ\vfndub{F}{1}{\circ}
        }.
    \end{align*}
\end{proof}

\end{appendices}
\end{document}

%% file: _math.tex
\usepackage{amsmath, amssymb, xcolor, bbm, amsthm, mathtools, adjustbox,
xcolor, booktabs, tcolorbox, enumitem, hyperref}
\usepackage[title]{appendix}

\newtheorem{theorem}{Theorem}[section]
\newtheorem{claim}{Claim}
\newtheorem{corollary}{Corollary}[theorem]
\newtheorem{lemma}[theorem]{Lemma}
\newtheorem{proposition}[theorem]{Proposition}
\newtheorem{definition}[theorem]{Definition}
\newtheorem{remark}{Remark}[theorem]

\newenvironment{subproof}[1][\proofname]{
    
    \begin{proof}[#1]
        }{
    \end{proof}
    }

\let\originalleft\left 
\let\originalright\right
\renewcommand{\left}{\mathopen{}\mathclose\bgroup\originalleft}
\renewcommand{\right}{\aftergroup\egroup\originalright}

\newcommand{\comma}{,\,}
\newcommand{\cm}{,\,}

\newcommand{\rvar}[1]{#1} 
\newcommand{\rvctr}[1]{\mathbf{#1}} 

\newcommand{\rvarsub}[2]{{#1}_{#2}} 
\newcommand{\rvctrsub}[2]{\mathbf{#1}_{#2}} 


\newcommand{\rvctrdub}[3]{\mathbf{#1}_{#2}^{#3}} 

\newcommand{\E}{\mathbb{E}\,} 
\newcommand{\Ewa}[1]{\mathbb{E}\,\left[{#1}\right]} 
\newcommand{\Esub}[1]{\mathbb{E}_{#1}\,} 
\newcommand{\Ewasub}[2]{\mathbb{E}_{#2}\,\left[{#1}\right]} 

\newcommand{\paren}[1]{\left({#1}\right)} 

\newcommand{\normsub}[2]{\left\lVert{#1}\right\rVert_{#2}}

\newcommand{\vctr}[1]{\mathbf{#1}} 

\newcommand{\vctrsup}[2]{\mathbf{#1}^{#2}} 
\newcommand{\mtrxsup}[2]{\mathbf{#1}^{#2}} 

\newcommand{\vctrsub}[2]{\mathbf{#1}_{#2}} 
\newcommand{\mtrxsub}[2]{\mathbf{#1}_{#2}} 

\newcommand{\vctrdub}[3]{\mathbf{#1}_{#2}^{#3}} 

\newcommand{\spc}[1]{\mathbb{#1}} 

\newcommand{\spcsub}[2]{\mathbb{#1}_{#2}} 

\newcommand{\spcsup}[2]{\mathbb{#1}^{#2}} 

\newcommand{\spcdub}[3]{\mathbb{#1}_{#2}^{#3}} 

\newcommand{\ind}[1]{\mathbbm{1}_{\left\{{#1}\right\}}} 


\newcommand{\fn}[1]{#1} 
\newcommand{\fnsub}[2]{{#1}_{#2}} 
\newcommand{\fnsup}[2]{{#1}^{#2}} 
\newcommand{\fndub}[3]{{#1}_{#2}^{#3}} 

\newcommand{\vfn}[1]{\mathbf{#1}} 
\newcommand{\vfnsub}[2]{\mathbf{#1}_{#2}} 
\newcommand{\vfnsup}[2]{\mathbf{#1}^{#2}} 
\newcommand{\vfndub}[3]{\mathbf{#1}_{#2}^{#3}} 

\newcommand{\fnwa}[2]{{#1}\left({#2}\right)} 
\newcommand{\fnwasub}[3]{{#1}_{#3}\left({#2}\right)} 
\newcommand{\fnwasup}[3]{{#1}^{#3}\left({#2}\right)} 
\newcommand{\fnwadub}[4]{{#1}_{#3}^{#4}\left({#2}\right)} 

\newcommand{\vfnwa}[2]{\mathbf{#1}\left({#2}\right)} 
\newcommand{\vfnwasub}[3]{\mathbf{#1}_{#3}\left({#2}\right)} 
\newcommand{\vfnwasup}[3]{\mathbf{#1}^{#3}\left({#2}\right)} 
\newcommand{\vfnwadub}[4]{\mathbf{#1}_{#3}^{#4}\left({#2}\right)} 

\newcommand{\ith}[1]{{#1}\textsuperscript{th}} 

\DeclareMathOperator*{\argmax}{arg\,max} 
\DeclareMathOperator*{\argmin}{arg\,min}

\DeclareMathOperator*{\st}{\text{ s.t. }}


\newcommand{\innersub}[3]{\left\langle {#1}, \, {#2}\right\rangle_{#3}} 

\newcommand{\dotpd}[2]{{#1}^\top{#2}} 

\newcommand{\kernel}{k} 

\newcommand{\kernelwa}[2]{k\left({#1}, \, {#2}\right)} 
\newcommand{\kernelwasub}[3]{k_{#3}\left({#1}, \, {#2}\right)} 

\newcommand{\gcpxwasub}[2]{\mathcal{G}_{#2}\left({#1}\right)} 
\newcommand{\egcpxwasub}[2]{\hat{\mathcal{G}}_{#2}\left({#1}\right)} 

\newcommand{\gcpxwadub}[3]{\mathcal{G}_{#2}^{#3}\left({#1}\right)} 
\newcommand{\egcpxwadub}[3]{\hat{\mathcal{G}}_{#2}^{#3}\left({#1}\right)} 

\newcommand{\seqdub}[3]{\left\{{#1}\right\}_{#2}^{#3}} 

\newcommand{\absolute}[1]{\left|{#1}\right|}
\newcommand{\cardin}[1]{\left|{#1}\right|} 

\newcommand{\lorc}[2]{\left({#1},\, {#2}\right]} 
\newcommand{\lcrc}[2]{\left[{#1},\, {#2}\right]} 


\newcommand{\boldalpha}{\boldsymbol{\alpha}}

\newcommand{\boldxi}{\boldsymbol{\xi}}

\newcommand{\setonly}[1]{\left\{{#1}\right\}} 
\newcommand{\setwvert}[2]{\left\{{#1} \,\vert\, {#2}\right\}} 
\newcommand{\setwcolon}[2]{\left\{{#1} \,\colon\, {#2}\right\}} 




\newcommand{\rmk}[1]{
    \begin{remark}
        {#1}
    \end{remark}
    }